\documentclass[10pt,twocolumn,letterpaper]{article}

\usepackage[pagenumbers]{cvpr} 

\usepackage{graphicx}
\usepackage{amsmath}
\usepackage{amssymb}
\usepackage{booktabs}
\usepackage[table]{xcolor}
\usepackage{bm}
\usepackage{float}
\usepackage{multirow}
\usepackage{balance}

\usepackage[pagebackref,breaklinks,colorlinks]{hyperref}

\usepackage[capitalize]{cleveref}
\crefname{section}{Sec.}{Secs.}
\Crefname{section}{Section}{Sections}
\Crefname{table}{Table}{Tables}
\crefname{table}{Tab.}{Tabs.}

\def\cvprPaperID{11967} 
\def\confName{CVPR}
\def\confYear{2023}

\newcommand{\caltech}{Caltech256}
\newcommand{\cifar}{Cifar10}
\newcommand{\cifarh}{Cifar100}
\newcommand{\cub}{CUB200}
\newcommand{\daircraft}{Decathlon Aircraft}
\newcommand{\ddtd}{Decathlon DTD}
\newcommand{\dflowers}{Decathlon Flowers}
\newcommand{\ducf}{Decathlon UCF101}
\newcommand{\eurosat}{EuroSAT}
\newcommand{\fgvcaircrafts}{FGVC Aircrafts}
\newcommand{\icassava}{iCassava}
\newcommand{\mitd}{MIT-67}
\newcommand{\oxfordflowers}{Oxford Flowers}
\newcommand{\pets}{Oxford Pets}
\newcommand{\stanfordcars}{Stanford Cars}
\newcommand{\stanforddogs}{Stanford Dogs}

\newcommand{\cityscapes}{Cityscapes}
\newcommand{\comic}{Comic}
\newcommand{\crowdhuman}{CrowdHuman}
\newcommand{\duo}{DUO}
\newcommand{\kitti}{KITTI}
\newcommand{\minneapple}{MinneApple}
\newcommand{\sixray}{SIXray}
\newcommand{\tabled}{table-detection}
\newcommand{\visdrone}{VisDrone}
\newcommand{\watercolor}{Watercolor}

\begin{document}

\title{A Meta-Learning Approach to Predicting Performance and Data Requirements}

\author{Achin Jain$^1$\thanks{Corresponding author. $^\dagger$ Work done at AWS.} \and 
Gurumurthy Swaminathan$^1$ \and
Paolo Favaro$^1$ \and
Hao Yang$^1$ \and
Avinash Ravichandran$^1$ \and
Hrayr Harutyunyan$^{1,2\dagger}$ \and
Alessandro Achille$^1$ \and
Onkar Dabeer$^1$ \and
Bernt Schiele$^1$ \and
Ashwin Swaminathan$^1$ \and
Stefano Soatto$^1$ \and
$^1$ AWS AI Labs, $^2$ University of Southern California \\
{\tt\small \{achij,gurumurs,pffavaro,haoyng,ravinash,aachille,onkardab,bschiel,swashwin,soattos\}@amazon.com}
}
\maketitle

\begin{abstract}
We propose an approach to estimate the number of samples required for a model to reach a target performance.
We find that the power law, the de facto principle to estimate model performance, leads to large error when using a small dataset (e.g., 5 samples per class) for extrapolation.
This is because the log-performance error against the log-dataset size follows a nonlinear progression in the few-shot regime followed by a linear progression in the high-shot regime.
We introduce a novel piecewise power law (PPL) that handles the two data regimes differently.
To estimate the parameters of the PPL, we introduce a random forest regressor trained via meta learning that generalizes across classification/detection tasks,  ResNet/ViT based architectures, and random/pre-trained initializations.
The PPL improves the performance estimation on average by 37\% across 16 classification and 33\% across 10 detection datasets, compared to the power law.
We further extend the PPL to provide a confidence bound and use it to limit the prediction horizon that reduces over-estimation of data by 76\% on classification and 91\% on detection datasets.
\end{abstract}

\section{Introduction}
\label{S:intro}
\begin{figure}[t!]
\small
\centering
\includegraphics[width=1\columnwidth]{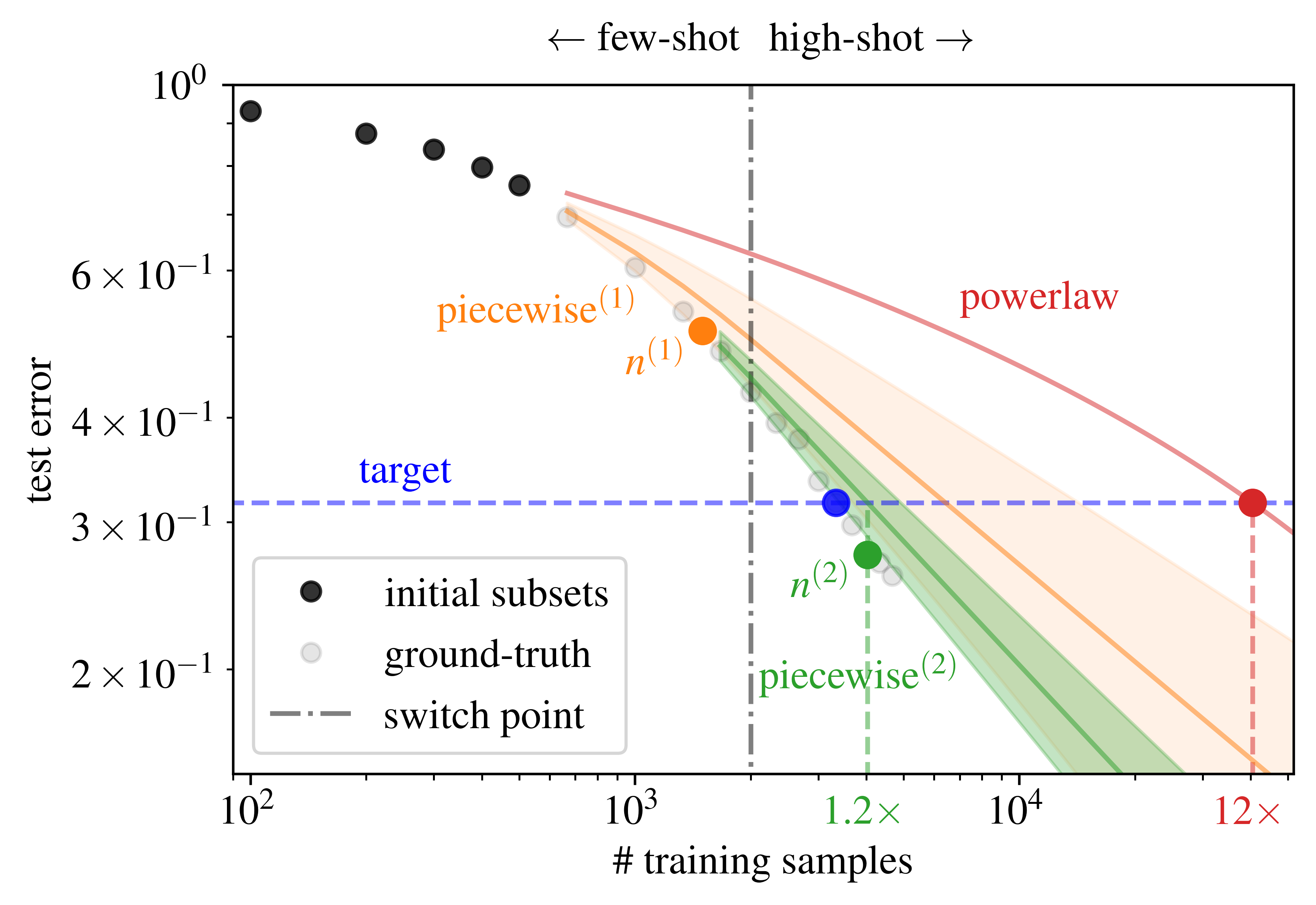}
\caption{Performance estimation curves using the powerlaw and piecewise power law (PPL) with estimated confidence. The PPL reduces over-estimation of the power law from 12$\times$ to 1.9$\times$ in 1 step, and further to 1.2$\times$ in 2 steps using the estimated confidence bounds to limit the prediction horizon $n^{(1)}$ in the first step.}
\label{F:main}
\vspace{-15pt}
\end{figure}

More data translates to better performance, on average, and higher cost.
As data requirements scale, there is a natural desire to predict the cost to train a model and what performance it may achieve, as a function of cost, without training.
Towards this goal, neural scaling laws~\cite{cortes1993learning,rosenfeld2020constructive,hestness2017deep,beery2018recognition,bahri2021explaining,hoiem2021learning} have been proposed that 
suggest that the performance of a model trained on a given dataset size follows a linear fit when 
plotted on a logarithmic scale (power law in linear scale).

In practice, however, the power law provides only a family of functions and its parameters must be fitted to each specific case 
for performance prediction. A common use case is one where a \textit{small} initial dataset is made available and can be used to obtain small-scale performance statistics 
that are relatively inexpensive to obtain and can be used to fit the power law parameters. 
Then, the fitted function is used to predict the performance for any dataset size training through extrapolation.
This approach is found empirically to generalize across several datasets and deep learning models \cite{rosenfeld2020constructive}. 
However, most applications of power law are done in the high-shot regime.
The few-shot regime (e.g., 5 samples/class) is particularly useful given the prevalence of pre-trained initializations available for model training.
In the few-shot regime, the performance curve exhibits a non-linear progression followed by a linear progression, see Figure~\ref{F:main}.
Thus, data requirements based on the power law can lead to significant errors incurring additional cost for acquiring data.

In this paper, we propose a piecewise power law (PPL) that models the performance as a quadratic curve in the few-shot regime and a linear curve in the high-shot regime in the log-log domain, while ensuring continuity during the transition.
To estimate the parameters of the PPL, we first identify the switching point between the quadratic and linear curves using PowerRF, a random forest regressor trained via meta-learning, and then use the performance statistics to fit the remaining parameters.
We show that our approach provides a better estimation of performance than the power law across several datasets, architectures, and initialization settings.
We extend the PPL to provide a confidence estimate that is used to further reduce the error in data estimation.
In Figure~\ref{F:main}, the confidence estimate controls the maximum number of samples in a step such that the PPL uses two steps to achieve the target performance with 1.2$\times$ over-estimation compared to 12$\times$ with the power law.

Our contributions are as follows.
We propose an improved policy for predicting data size needed to achieve a target accuracy with three main innovations: (1) a piecewise power law model (PPL) that approximates the performance error curve from low-to-high-shot regime, (2) a meta-learning approach to estimate the parameters of the PPL, and (3) incorporating the confidence interval of the estimates to limit the prediction horizon and reduce error in data estimation.
We demonstrate the generalization of the PPL and the meta-model on 16 classification and 10 detection datasets, improving the (1) performance estimates by 37\% on classification and 33\% on detection datasets and (2) data estimates by 76\% on classification and 91\% on detection datasets, with respect to the power law.

\section{Related work}
\label{S:prior_work}

The power law is considered as the \textit{de facto} principle for modeling learning curves~\cite{cortes1993learning,mahmood2022much,rosenfeld2020constructive,bahri2021explaining,hestness2017deep,hoiem2021learning,abnar2022exploring}.
It has also been used for analysis and evaluation of design choices such as pre-training, architecture, and data augmentation~\cite{hoiem2021learning}.
The focus of our work, however, is not just on evaluating the fit quality of a predictor 
on the learning curves but also using it to predict performance through extrapolation and estimate data requirements to reach a target performance.
In one of the closely related work, Mahmood et al.~\cite{mahmood2022much} evaluate the power law along with other modeling techniques such as algebraic, arctan, and logarithmic in predicting performance and data requirements.
They improve the estimates of data requirements via a correction factor and multiple rounds of data collection.
In a follow up and \textit{concurrent} work, Mahmood et al.~\cite{mahmood2022optimizing} propose stochastic optimization with the power law as a predictor to estimate data requirements.
In contrast to these works, this paper focuses on the few-shot setting (e.g., when an initial dataset has 5 samples per class) where the power law breaks as the log-performance versus log-dataset sizes follows a non-linear progression.
We not only propose an extension to the power law by modeling the few-shot and high-shot regimes differently but also introduce a robust strategy that uses model confidence to reduce data estimation errors over multiple rounds.

Several works~\cite{hestness2017deep,rosenfeld2020constructive,caballero2022broken,alabdulmohsin2022revisiting} have shown that the learning curves exhibit multiple regimes.
Hestness et al.~\cite{hestness2017deep} argue about the existence of three data regimes: small data, power law, and irreducible error.
Rosenfeld et al.~\cite{rosenfeld2020constructive} model the transition from the small data regime to the power law regime using a complex envelope function.
However, unlike this paper, they mainly focus on the fit quality across different model and data configurations, and the extrapolation results on few-shot datasets show poor generalization.
In a \textit{concurrent} work, Caballero et al.~\cite{caballero2022broken} propose the broken neural scaling law, which is a smoothly interpolated piecewise linear curve to model different regimes such as initial random fitting, power law, double descent, and saturation.
Alabdulmohsin et al.~\cite{alabdulmohsin2022revisiting} also model the learning curves in multiple regimes.
In \cite{caballero2022broken,alabdulmohsin2022revisiting}, the setting is different than ours in that (1) a predictor is fit on subsets of JFT300~\cite{JFT300} used for pretraining with evaluation on the downstream datasets, (2) the primary focus is on evaluating the curve fitting and not on estimating data requirements, and (3) the  evaluation is done in the very large data regime with millions of samples which is not encountered in real-life models because data is scarce and is usually not enough to saturate the performance of the models.
In this paper, we mainly focus on performance extrapolation with a few-shot dataset.

\section{Predicting performance and data requirements}
\label{S:method}

We consider the problem of extrapolating the performance (test error) of a trained model, had we trained it on a larger number of samples. This prediction is useful to estimate the data requirements to achieve a given performance target, a fundamental step in the deployment of a model in real-world applications.
Specifically, given a model $\mathcal{M}$ trained on a dataset $\mathcal{D}^{(0)}$ with $n^{(0)}$ training samples, we want to predict its performance on dataset sizes $n>n^{(0)}$.
We pose this problem in the practical use case of few-shot learning:
(1) $\mathcal{D}^{(0)}$ 
contains only a few samples per class, and (2) 
$\mathcal{M}$ is fine-tuned after pre-training on ImageNet~\cite{ImageNet} for classification tasks and COCO~\cite{COCO} for object detection.
We denote $v(n)$ as the true performance score of $\mathcal{M}$ obtained by fine-tuning and $\hat{v}(n)$ as the predictor of the performance of $\mathcal{M}$.
We choose Top-1 Accuracy for classification and mean Average Precision (mAP) for detection as the score metric.

\subsection{Problem statement}

\noindent\textbf{Predicting performance.}
For a given dataset, we construct $M$ subsets $\mathcal{S}_1 \subset \mathcal{S}_2 \subset ... \subset \mathcal{S}_{m} = \mathcal{D}^{(0)} \subset \mathcal{S}_{m+1} \subset ... \subset \mathcal{S}_{M} = \mathcal{D}^{(\text{FULL})} $.
Here $\mathcal{D}^{(0)}$ is the initial dataset we use for extrapolation in practice and $\mathcal{D}^{(\text{FULL})}$ is a larger dataset obtained by adding samples to $\mathcal{D}^{(0)}$.
We fine-tune $\mathcal{M}$ on each subset to obtain $\{n_{i}, v(n_{i})\}_{i=1}^M$ sample pairs, where $n_i=|\mathcal{S}_i|$ and $v(n_i)$ is the performance obtained by the model trained on dataset $\mathcal{S}_i$.
Then, we fit a predictor $\hat{v}(n)$ on samples $\{n_{i}, v(n_{i})\}_{i=1}^m$ and evaluate its extrapolation performance on $\{n_{i}, v(n_{i})\}_{i=m+1}^M$.
Our goal is to build a predictor that achieves 
the smallest
mean prediction error
\begin{align}
    \label{E:mpe}
     \mathcal{E}_{\mathrm{perf}} = \frac{1}{M-m} \sum_{i=m+1}^{M} \left| v(n_{i}) - \hat{v}(n_{i}) \right|.
\end{align}

\noindent\textbf{Predicting data requirements.}
Next, we estimate the number of samples that should be added to $\mathcal{D}^{(0)}$ to reach a target performance $v^*$.
To do so, we invert the predictor $\hat{v}(n)$ and get the first  estimate of the data requirements $n^{(1)} = \hat{v}^{-1}(v^*)$.
Due to modeling errors, the first estimate may over/under-estimate the number of samples $n^*\doteq v^{-1}(v^\ast)$ needed to reach the desired target.
In the case of over-estimation, i.e., $v(n^{(1)}) > v^*$ or equivalently $n^{(1)} > n^*$ , we collect 
further $n^{(1)}-n^{(0)}$ samples, where $n^{(0)}=|\mathcal{D}^{(0)}|$,
and stop.
In the case of under-estimation, it is desirable to update the predictor with the collected data and the corresponding model performance after re-training with the additional performance pair $\{n^{(1)}, v(n^{(1)})\}$, and to take additional steps 
to reach the target performance.
This is referred to as the \emph{data collection problem} in~\cite{mahmood2022much}.
We consider the same setting as~\cite{mahmood2022much}, where the maximum number of steps is limited to $T$ since each round of data collection and annotation can be costly to initiate.
Specifically, at the $k$-th step of the data collection, we fit the predictor $\hat{v}_k$ on $\{n_{i}, v(n_{i})\}_{i=1}^m \cup \{n^{(i)}, v(n^{(i)})\}_{i=1}^{k-1}$ to obtain the data requirement estimate $n^{(k)}$.
Our goal is to achieve a small data estimation error $|\mathcal{E}_{\mathrm{data}}|$ where
\small
\begin{gather}
    \label{E:dee}
    \mathcal{E}_{\mathrm{data}} = \frac{n^{(K)} - n^*}{n^*}, \\
    K = \min \Bigl\{ T, \min \{k: \hat{v}_k(n^{(k)})>v^* \} \Bigl\} \nonumber.
\end{gather}
\normalsize
$\mathcal{E}_{\mathrm{data}}<0$ represents an under-estimate and $\mathcal{E}_{\mathrm{data}}>0$ an over-estimate.

\noindent\textbf{Challenges in the few-shot setting.}
In the few-shot setting, the largest subset used to fit a predictor $\mathcal{S}_{m} = \mathcal{D}^{(0)}$ contains only a few samples per class (e.g., we consider 5 for classification and 10 for object detection).
We observe that the power law~\cite{cortes1993learning} as a predictor breaks down  in this setting when $n_{M} \gg n^{(0)}$, i.e., when we extrapolate its performance to the high-shot regime starting from the few-shot regime.
To this end, we propose a piecewise power law that models the two regimes differently.
In addition, a direct estimate $n^{(k)} = \hat{v}_k^{-1}(v^*)$ for predicting data requirements is not always the best choice, especially when the size of the initial dataset $n^{(0)}\ll n^*$, i.e, $\hat{v}_1$ is fit to the data samples from the few-shot regime and the target performance can be achieved only in the high-shot regime.
We propose a strategy that uses model confidence of the predictor to reduce over-estimation of data requirements, a problem that plagues the prior approach such as the power law.

\subsection{The piecewise power law (PPL)}
\label{SS:ppl}
We empirically observe that the test error in logarithmic scale $\log(1-v(n))$ varies nonlinearly as a function of $\log(n)$ in the few-shot regime and linearly in the high-shot regime, see an illustration in Figure~\ref{F:main}.
Also, the point of separation of these data regimes differs for every dataset.
Thus, we design the predictor $\hat{v}(n)$ to be smooth and monotonic, such as the power law, but also as a piecewise function that can differentiate between the two data regimes.
Additionally, to model the stochasticity in training models on the subsets, we design $\hat{v}(n)$ to predict both the mean performance and its uncertainty.
The uncertainty estimate is used to predict data requirements as outlined in the following section.

To model the nonlinear behavior in the few-shot regime, we consider a quadratic model.
Thus, the piecewise power law is defined as
\small
\begin{align}
    \small
    \label{E:ppl}
    \log(1-\hat{v}(n;\bm{\theta})) = 
        \begin{cases}
            \theta_1 + \theta_2 \log(n) + \theta_3 \log(n)^2,& \text{if } n\leq N\\
            \theta_4 + \theta_5 \log(n),              & \text{otherwise}
        \end{cases}
\end{align}
\normalsize
\noindent where $\{\theta_{i}\}_{i=1}^5$ are the coefficients of the model and $N$ is the switching point from a quadratic to a linear
function
that differentiates between the few-shot and the high-shot data regimes.
By enforcing continuity and differentiability at $N$, $\theta_4$ and $\theta_5$ 
become functions of $\bm{\theta}=[\theta_1, \theta_2, \theta_3]^T$ and $N$.
Thus, the piecewise power law \eqref{E:ppl} effectively has only 4 parameters.
Given the data samples $\{n_{i}, v(n_{i})\}_{i=1}^m$, we first obtain $N$ through a meta-model trained on a dictionary of learning curves of several datasets (see Section~\ref{SS:meta-learning} for details), and then derive $\bm{\theta}$ by fitting the piecewise model \eqref{E:ppl} on $\{n_{i}, v(n_{i})\}_{i=1}^m$ via nonlinear least squares in the logarithmic scale using the Levenberg–Marquardt algorithm on
\small
\begin{gather}
    \label{E:nls}
     \min_{\bm{\theta}} \ \|\mathbf{y}-\mathbf{\hat{y}}(\bm{\theta})\|^2 \\
    \mathbf{y}_i = y_i = \log(1-v(n_{i})), \quad \forall i \in \{1, 2, ..., m\}, \nonumber\\
    \mathbf{\hat{y}}_i = \hat{y}_i = \log(1-\hat{v}(n_{i})), \quad \forall i \in \{1, 2, ..., m\}. \nonumber
\end{gather}
\normalsize
We estimate the mean $\mu_{\hat{v}}(n)$ and the variance $\sigma^2_{\hat{v}}(n)$ (due to modeling errors) of the predictor as follows.
First, we compute of the covariance matrix of the parameters~\cite{Gavin2013TheLM} as
\small
\begin{gather}
    \bm{\Sigma}_{\bm{\theta}} = (\mathbf{J}^T\mathbf{J})^{-1}, \ \mathbf{J} = \left[\frac{\partial \mathbf{\hat{y}}(\bm{\theta})}{\partial\bm{\theta}}\right]_{m\times3}
\end{gather}
\normalsize
where $\mathbf{J}$ is the Jacobian matrix of $\mathbf{\hat{y}}$ with respect to $\bm{\theta}$.
By rewriting \eqref{E:ppl} as $\hat{y}(n; \bm{\theta}) = \bm{\alpha}^T(n) \bm{\theta}$, we obtain the variance $\sigma^2_{\hat{y}}(n) = \bm{\alpha}^T(n) \bm{\Sigma}_{\bm{\theta}} \bm{\alpha}(n)$, where $\bm{\alpha}(n)$ is given by
\small
\begin{gather}
    \bm{\alpha}(n) = 
    \begin{cases}
            [1, \log(n), \log(n)^2], & \text{if } n\leq N\\
            [1, \log(n), 2N\log(n)-N^2]  & \text{otherwise.}
        \end{cases}
\end{gather}
\normalsize
The latter condition for $n>N$ comes from the continuity and differentiability constraints at $N$.
Now, by assuming $\hat{y}$ is normally distributed, $\hat{v} = 1-\exp(\hat{y})$ is log-normally distributed with mean $\mu_{\hat{v}}$ and variance $\sigma^2_{\hat{v}}$ given by
\small
\begin{align}
    \mu_{\hat{v}}(n) &= 1 - \exp\left(\hat{y}(n) + \frac{\sigma^2_{\hat{y}}(n)}{2}\right), \\
    \sigma^2_{\hat{v}}(n) &= \exp\left(\hat{y}(n) + \frac{\sigma^2_{\hat{y}}(n)}{2}\right) \sqrt{\exp(\sigma^2_{\hat{y}}(n) - 1)}.
\end{align}
\normalsize

\noindent\textbf{Using confidence bounds to predict data requirements.}
The test error in logarithmic scale $\log(1-v(n))$ decays rather slowly in the few-shot regime and we observe that the power law~\cite{cortes1993learning} can lead to large over-estimation in data requirements, i.e., $ \mathcal{E}_{\mathrm{data}} \gg 0$; for example, the power law over-estimates by 12$\times$ in Figure~\ref{F:main}.
Increasing the number of steps in the data collection process is not helpful since $v(n^{(1)}) > v^*$ in the first step itself.
We propose a strategy that exploits the model confidence of the predictor to prevent large estimation errors.
Specifically, we limit the number of samples in a step such that $3\sigma_{\hat{v}_k}(n^{(k)}) \leq \tau$, where $\tau$ is the confidence threshold.
Under this policy, the number of samples in step $k$ of the data collection process is determined by
\begin{align}
    \label{E:ppl_step}
    n^{(k)} = \min \left \{\mu_{\hat{v}_k}^{-1}(v^*), \sigma_{\hat{v}_k}^{-1}(\tau/3) \right\}
\end{align}
where $\hat{v}_k$ is fit on $\{n_{i}, v(n_{i})\}_{i=1}^m \cup \{n^{(i)}, v(n^{(i)})\}_{i}^{k-1}$.
In Figure~\ref{F:main}, we first add $n^{(1)}$ samples using~\eqref{E:ppl_step} and then we achieve the target in the next step with $n^{(2)}$ samples while satisfying $3\sigma_{\hat{v}_2}(n^{(2)}) \leq \tau$.
We empirically show that the same $\tau$ works consistently for all datasets across classification and object detection tasks.

\noindent\textbf{Comparison with the power law.}
The power law~\cite{cortes1993learning} predictor of model performance is given by
\begin{align}
    \label{E:powerlaw}
    1-\hat{v}(n;\bm{\theta}) = \theta_1 n^{\theta_2} + \theta_3.
\end{align}
The model comprises of 3 parameters $\bm{\theta}=[\theta_1,\theta_2,\theta_3]$.
The special case of the piecewise power law~\eqref{E:ppl} with $N=0$ is equivalent to the power law~\eqref{E:powerlaw} when $\theta_3=0$, that is
\begin{align}
    \label{E:powerlaw2}
    \log(1-\hat{v}(n;\bm{\theta})) = \theta_1 + \theta_2 \log(n),
\end{align}
whose parameters can simply be obtained by solving linear regression in the logarithmic space.
The parameter $\theta_3$ models the asymptotic error of the learning curves.
For most datasets we consider in the few-shot setting (with pre-training), we empirically observe that the predictor~\eqref{E:powerlaw2} achieves comparable or smaller $\mathcal{E}_{\mathrm{perf}}$ than predictor~\eqref{E:powerlaw}; see Appendix~\ref{A:powerlaw_approx}.

\subsection{Meta-learning for the piecewise power law}
\label{SS:meta-learning}
We need to estimate four parameters in the piecewise power law~\eqref{E:ppl}: $\bm{\theta} \in \mathbb{R}^3$ and $N$ by fitting on $\{n_{i}, v(n_{i})\}_{i=1}^m$. $N$ is determined by the transition point between the non-linear and the linear regime. However, identifying an optimal $N$ is non-trivial given that many choices of $N$ can produce fit with low errors.
We find that a brute-force search that chooses $N$ based on fitting $\{n_{i}, v(n_{i})\}_{i=1}^{m-1}$ and evaluating on $\{n_{m}, v(n_{m})\}$ is better than the power law in several cases, but still far from the upper bound performance given by the piecewise power law.

To bridge the gap, we propose to learn a meta-model that, given data samples $\{n_{i}, v(n_{i})\}_{i=1}^m$, leverages knowledge from a dictionary of learning curves to predict the switching point $N$ between the quadratic and the linear functions.
Specifically, we train a Random Forest~\cite{breiman2001random} regressor $\mathcal{F}$ via meta-learning to predict
\small
\begin{align}
    \hat{N} = \mathcal{F}
    \left(
    \{ \log(n_{i}), \log(v(n_{i})) \}_{i=1}^m, \log(C)
    \right),
\end{align}
\normalsize
where $C$ is the number of classes in a dataset.
Once the meta-model predicts $\hat{N}$, other parameters of the piecewise power law are determined by fitting on the subsets of performance pairs with dataset size either smaller or larger than $\hat{N}$, via the optimization of Eq.~\eqref{E:nls} as described in Section~\ref{SS:ppl}.
We choose to predict only the parameter $\hat{N}$ using the meta-model as this can be learnt from the statistics of the learning curves, and the other parameters are better estimated from the data samples $\{n_{i}, v(n_{i})\}_{i=1}^m$.
We demonstrate in Section~\ref{SS:ablation} that the PPL is robust to errors in $N$.

\noindent\textbf{Training and inference using the meta-model.}
To train the meta-model, we use data samples $\{n_{i}, v(n_{i})\}_{i=1}^M$ from all subsets of \textit{both} few-shot and high-shot regimes.
For discrete choices of $N \in \{n_{1}, n_{2},\dots,n_{M}\}$, we fit the piecewise power law eq.~\eqref{E:ppl} on samples from the few-shot region $\{n_{i}, v(n_{i})\}_{i=1}^m$ and compute the mean prediction error $\mathcal{E}_{\mathrm{perf}}$ as defined in eq.~\eqref{E:mpe}.
The $N$ that minimizes the evaluation error is considered to be the ground-truth $N^*$ for the dataset.
We compute $N^*$ for several datasets and train the meta-model in a leave-one-out (LOO) fashion, i.e., during inference the meta-model used to evaluate on dataset $\mathcal{D}$ is trained on datasets \textit{excluding} $\mathcal{D}$.
In our experiments, we show generalization of the meta-model to several datasets, model architectures, and training settings.
Lastly, due to very distinct numbers of object categories in the datasets, we train a separate set of meta-models for classification and object detection tasks.

\section{Experiments}
\noindent\textbf{Datasets.}
We evaluate the piecewise power law on 16 classification and 10 detection datasets from diverse domains.
The datasets vary widely in terms of complexity measured by number of classes and number of training images.
Refer to the dataset statistics in Appendix~\ref{A:datasets}.

\noindent\textbf{Subsets used for fitting and evaluation.}
To provide a comprehensive analysis, we consider different subsets for fitting and evaluation of the performance predictors.

The \textbf{few-shot setting}
is the main focus of this paper and is most widely used across all experiments.
For most classification datasets with the exceptions listed below, we fit a performance predictor on the 5 subsets comprising of \{1, 2, 3, 4, 5\} samples per class.
For \cifar \ and \eurosat, the subsets for fitting comprise of \{5, 10, 15, 20, 25\} samples per class, and for \icassava\  \{10, 15, 20, 25, 30\} samples per class to ensure the number of training images in the smallest subset is more than the batch size of 32 used universally across all classification experiments.
For all classification datasets, we compute $\mathcal{E}_{\mathrm{perf}}$~\eqref{E:mpe} on the subsets comprising of \{10, 15, ..., 100\}\% of the dataset.
For all object detection datasets, we fit a performance predictor on the 5 subsets comprising of \{1, 5, 10, 15, 20\} samples per class and compute $\mathcal{E}_{\mathrm{perf}}$~\eqref{E:mpe} on the subsets comprising of \{25, 30, ..., 100\} samples per class.
To construct the subsets, we follow the natural k-shot sampling protocol~\cite{lee2022rethinking}.
Due to high variance in the training statistics, we choose a larger number of samples per class for fitting for the detection tasks compared to the classification tasks.
We also evaluate our approach in the \textbf{mid-shot setting} for the classification datasets where a predictor is (1) fit on \{10, 15, 20, 25, 30\}\% data and evaluated on \{35, 40, ..., 100\}\% data, and (2) fit on \{10, 20, 30, 40, 50\}\% data and evaluated on \{55, 60, ..., 100\}\% data.

We repeat all experiments with three different random seeds selecting different set of images for fitting and evaluation.
For training recipe used for finetuning and linear probing the subsets, see Appendix~\ref{A:implementation}.

\subsection{Extrapolation from few-shot to high-shot}
\label{SS:extrapolation_fewshot}
\begin{table}[t]
\rowcolors{2}{gray!10}{white}
\small
\centering
\caption{Mean prediction error $\mathcal{E}_{\mathrm{perf}}$~\eqref{E:mpe} for extrapolating performance from few-shot to high-shot. Piecewise GT denotes the upper bound. Piecewise outperforms powerlaw on 12/16 datasets. 
}
\label{T:performance_fewshot_classification}
\footnotesize
\resizebox{1.02\columnwidth}{!}{
\begin{tabular}{lrrrr}
\toprule
\multicolumn{5}{c}{CLASSIFICATION}                                                                                                                                       \\
\midrule
                    & \multicolumn{1}{c}{\textbf{powerlaw}} & \multicolumn{1}{c}{\textbf{arctan}} & \multicolumn{1}{c}{\textbf{piecewise}} & \multicolumn{1}{c}{\textbf{piecewise}} \\
\rowcolor{white}                    
                    & \multicolumn{1}{c}{\cite{cortes1993learning}}                  & \multicolumn{1}{c}{\cite{mahmood2022much}}                & \multicolumn{1}{c}{\textbf{(ours)}}    & \multicolumn{1}{c}{\textbf{(GT)}}      \\  
\midrule                    
\caltech~\cite{griffin2019caltech}          & 10.3±3.6                              & 3.0±1.5          & \textbf{2.0±0.9}   & 1.2±0.5               \\
\cifar~\cite{krizhevsky2009learning}             & 6.7±2.2                               & 6.0±6.2          & \textbf{0.9±0.5}   & 0.5±0.4               \\
\cifarh~\cite{krizhevsky2009learning}            & 6.5±3.1                               & 17.2±7.3         & \textbf{6.1±3.5}   & 5.3±3.8               \\
\cub~\cite{WelinderEtal2010}              & \textbf{2.6±0.3}                      & 14.1±3.9         & 4.0±0.1            & 0.7±0.1               \\
\daircraft~\cite{maji13fine-grained}\hspace{-5pt} & 18.0±1.8                              & 23.5±15.9        & \textbf{11.1±4.2}  & 11.1±4.1              \\
\ddtd~\cite{cimpoi2014describing}      & \textbf{3.2±1.9}                      & 4.7±2.5          & 5.6±1.7            & 2.1±1.1               \\
\dflowers~\cite{Nilsback06}\hspace{-5pt}  & \textbf{1.0±0.3}                      & 1.5±0.3          & 2.0±0.3            & 1.1±0.0               \\
\ducf~\cite{soomro2012dataset}\hspace{-5pt}   & 14.5±1.9                              & 15.5±4.9         & \textbf{4.1±3.0}   & 4.1±2.9               \\
\eurosat~\cite{helber2018introducing,helber2019eurosat}             & 2.6±0.6                               & 4.2±2.4          & \textbf{0.9±0.2}   & 0.9±0.2               \\
\fgvcaircrafts~\cite{maji13fine-grained}     & 25.8±1.6                              & \textbf{9.2±7.4} & 19.1±1.4           & 11.1±1.8              \\
\icassava~\cite{mwebaze2019icassava}            & 9.2±6.7                               & 14.6±4.8         & \textbf{6.9±2.4}   & 6.9±2.4               \\
\mitd~\cite{quattoni2009recognizing}               & \textbf{4.2±1.7}                      & 8.2±5.3          & 4.3±2.5            & 3.9±2.3               \\
\oxfordflowers~\cite{Nilsback06}       & 1.5±0.4                               & 1.5±0.3          & \textbf{1.2±0.3}   & 1.1±0.4               \\
\pets~\cite{parkhi12a}                & 9.2±0.4                               & \textbf{1.1±0.5} & 5.6±0.8            & 1.7±0.5               \\
\stanfordcars~\cite{KrauseStarkDengFei-Fei_3DRR2013}        & 26.4±1.3                              & 17.6±2.9         & \textbf{17.3±2.7}  & 7.7±3.3               \\
\stanforddogs~\cite{KhoslaYaoJayadevaprakashFeiFei_FGVC2011}        & 6.1±5.4                               & 7.8±2.7          & \textbf{2.3±1.0}   & 1.2±0.1               \\
\midrule
AVERAGE             & 9.2±2.1                               & 9.3±4.3          & \textbf{5.8±1.6}   & 3.8±1.5                \\
\bottomrule
\end{tabular}
}
\end{table}

We first evaluate the piecewise model in the few-shot setting with the mean prediction error $\mathcal{E}_{\mathrm{perf}}$~\eqref{E:mpe} as the metric for evaluation.
We compare the piecewise power law~\eqref{E:ppl} with the power law~\cite{cortes1993learning,mahmood2022much} on classification tasks in Table~\ref{T:performance_fewshot_classification} and object detection tasks Table~\ref{T:performance_fewshot_detection}.
Other predictors such as algebraic~\cite{mahmood2022much}, arctan~\cite{mahmood2022much}, and logarithmic~\cite{mahmood2022much} did not perform better in our experiments; we compare with ``arctan'' here, see others in Appendix~\ref{A:extrapolation_fewshot}.
The upper bound performance of the piecewise power law is denoted by ``piecewise GT''.
This corresponds to the piecewise model using the ground-truth switching point $N^*$ as described in Section~\ref{SS:meta-learning}.
Note that the upper bound is not always close to zero because the derivation of $N^*$ utilizes only five data samples from the few-shot regime for fitting the piecewise power law.
The ``piecewise'' power law with the meta-model performs better on 12/16 classification tasks and 9/10 object detection tasks reducing the average mean prediction error of the ``powerlaw'' by 37\% and 33\%, respectively.
It also achieves lower variance consistently across most datasets.
We note that there is still a gap between the performance of the meta-model with respect to the upper bound.

\subsection{Extrapolation from mid-shot to high-shot}
\begin{table}[t!]
\rowcolors{2}{gray!10}{white}
\small
\centering
\caption{Mean prediction error $\mathcal{E}_{\mathrm{perf}}$~\eqref{E:mpe} for extrapolating performance from few-shot to high-shot. Piecewise GT denotes the upper bound. Piecewise outperforms powerlaw on 9/10 datasets.}
\label{T:performance_fewshot_detection}
\footnotesize
\resizebox{\columnwidth}{!}{
\begin{tabular}{lrrrr}
\toprule
\multicolumn{5}{c}{DETECTION}                                                                                                                                           \\
\midrule
                    & \multicolumn{1}{c}{\textbf{powerlaw}} & \multicolumn{1}{c}{\textbf{arctan}} & \multicolumn{1}{c}{\textbf{piecewise}} & \multicolumn{1}{c}{\textbf{piecewise}} \\
\rowcolor{white}                    
                    & \multicolumn{1}{c}{\cite{cortes1993learning}}                  & \multicolumn{1}{c}{\cite{mahmood2022much}}                & \multicolumn{1}{c}{\textbf{(ours)}}    & \multicolumn{1}{c}{\textbf{(GT)}}      \\  
\midrule
\cityscapes~\cite{cityscapes}   & 1.3±0.8                               & 1.5±0.6          & \textbf{1.1±0.5}   & 0.9±0.6               \\
\comic~\cite{cartoon}        & 4.4±2.9                               & 28.0±33.9        & \textbf{4.1±1.0}   & 3.4±2.0               \\
\crowdhuman~\cite{crowdhuman}   & 0.8±0.2                               & 1.5±0.5          & \textbf{0.7±0.3}   & 0.5±0.3               \\
\duo~\cite{duo}          & 3.9±1.8                               & 4.5±1.6          & \textbf{2.4±0.5}   & 1.8±0.9               \\
\kitti~\cite{kitti}        & 2.6±2.1                               & \textbf{1.5±1.0} & 1.6±0.7            & 1.5±1.4               \\
\minneapple~\cite{minneapple}   & 4.7±2.3                               & \textbf{1.1±0.3} & 1.9±1.0            & 0.6±0.1               \\
\sixray~\cite{sixray}       & 6.9±0.9                               & 8.3±9.9          & \textbf{2.7±2.7}   & 2.4±1.1               \\
\tabled~\cite{table}        & 5.9±2.7                               & 7.8±2.2          & \textbf{5.5±0.8}   & 5.5±2.2               \\
\visdrone~\cite{visdrone} & \textbf{0.3±0.1}                      & 0.7±0.3          & 0.8±0.3            & 0.4±0.1               \\
\watercolor~\cite{cartoon}   & 5.2±1.5                               & 6.7±2.7          & \textbf{3.2±1.4}   & 3.1±1.7               \\
\midrule
AVERAGE      & 3.6±1.5                               & 6.2±5.3          & \textbf{2.4±0.9}   & 2.0±1.0                \\
\bottomrule
\end{tabular}
}
\end{table}

\begin{table}[t!]
\rowcolors{2}{gray!10}{white}
\centering
\caption{Mean prediction error $\mathcal{E}_{\mathrm{perf}}$~\eqref{E:mpe} for extrapolating performance from mid-shot to high-shot. Piecewise works better on 14/16 and 12/16 datasets using 30\% and 50\% data, respectively.}
\label{T:performance_midshot}
\resizebox{\columnwidth}{!}{
\begin{tabular}{lrrrr}
\toprule
\multicolumn{5}{c}{CLASSIFICATION}                                                                                                                                       \\
\midrule
                    & \multicolumn{2}{c}{\textbf{using 30\% data}}                                   & \multicolumn{2}{c}{\textbf{using 50\% data}}                                   \\
                    & \multicolumn{1}{c}{\textbf{powerlaw}} & \multicolumn{1}{c}{\textbf{piecewise}} & \multicolumn{1}{c}{\textbf{powerlaw}} & \multicolumn{1}{c}{\textbf{piecewise}} \\
\midrule                    
\caltech          & 0.8±0.6                               & \textbf{0.6±0.3}                       & 0.5±0.2                               & \textbf{0.3±0.0}                       \\
\cifar             & 0.3±0.3                               & \textbf{0.1±0.0}                       & 0.3±0.1                               & \textbf{0.1±0.0}                       \\
\cifarh            & 1.1±0.8                               & \textbf{0.6±0.1}                       & \textbf{0.3±0.1}                      & \textbf{0.3±0.1}                       \\
\cub              & 4.5±2.4                               & \textbf{2.5±1.3}                       & 1.6±0.6                               & \textbf{0.9±0.0}                       \\
\daircraft & 8.5±1.6                               & \textbf{4.1±2.0}                       & 4.2±0.3                               & \textbf{1.5±1.1}                       \\
\ddtd      & \textbf{1.2±0.4}                      & 1.7±0.8                                & 1.4±0.8                               & \textbf{1.3±0.4}                       \\
\dflowers  & \textbf{1.4±0.4}                      & 2.6±1.7                                & \textbf{1.0±0.3}                      & 1.8±0.3                                \\
\ducf   & 2.7±1.5                               & \textbf{2.2±1.2}                       & 1.0±0.4                               & \textbf{0.8±0.3}                       \\
\eurosat             & 0.3±0.2                               & \textbf{0.1±0.0}                       & 0.2±0.1                               & \textbf{0.1±0.0}                       \\
\fgvcaircrafts     & 6.4±4.7                               & \textbf{2.0±0.8}                       & 2.6±0.9                               & \textbf{0.9±0.4}                       \\
\icassava            & 2.4±0.7                               & \textbf{1.2±0.5}                       & 0.5±0.3                               & \textbf{0.5±0.1}                       \\
\mitd               & 2.3±1.3                               & \textbf{1.1±0.4}                       & 1.2±0.7                               & \textbf{0.9±0.5}                       \\
\oxfordflowers       & 3.6±1.8                               & \textbf{1.9±0.7}                       & \textbf{0.6±0.3}                      & 0.7±0.5                                \\
\pets                & 2.2±0.9                               & \textbf{1.1±0.2}                       & 1.2±0.8                               & \textbf{0.7±0.4}                       \\
\stanfordcars        & 9.7±0.1                               & \textbf{1.1±0.4}                       & 4.3±0.9                               & \textbf{0.4±0.1}                       \\
\stanforddogs        & 3.1±2.0                               & \textbf{2.1±0.3}                       & \textbf{1.2±1.2}                      & 1.6±0.3                                \\
\midrule
AVERAGE                & 3.2±1.2                               & \textbf{1.6±0.7}                       & 1.4±0.5                               & \textbf{0.8±0.3}                      \\
\bottomrule
\end{tabular}
}
\end{table}
\begin{table*}[t!]
\caption{Left: Data estimation error $\mathcal{E}_{\mathrm{data}}$ to reach the target performance corresponding to 90\% samples (of the full dataset). $\mathcal{E}_{\mathrm{data}}<0$ represents an under-estimate and $\mathcal{E}_{\mathrm{data}}>0$ an over-estimate.
PPL  with confidence threshold (5\%) achieves the lowest error on 13/16 classification and 9/10 detection tasks. Right: PPL with confidence (5-step) achieves smaller $\mathcal{E}_{\mathrm{data}}$ (closer to 0) than powerlaw (5-step) for different performance targets obtained by using \{50, 60, 70, 80, 90\}\% of the full dataset (more datasets in Appendix~\ref{A:data_fewshot}).
}
\begin{minipage}{0.67\textwidth}
    \rowcolors{2}{gray!10}{white}
    \footnotesize
    \centering
    \resizebox{1\textwidth}{!}{
    \begin{tabular}{l|rr|rrrr}
    \toprule
    \multicolumn{7}{c}{CLASSIFICATION}                                                                                                                                                                       \\
    \midrule
                        & \multicolumn{2}{c}{\textbf{powerlaw}~\cite{cortes1993learning}}                            & \multicolumn{4}{c}{\textbf{piecewise (ours)}}                                                                                            \\
    \cmidrule(lr){2-3} \cmidrule(lr){4-7}                    
                        & \multicolumn{1}{c}{1-step} & \multicolumn{1}{c}{5-step} & \multicolumn{1}{c}{1-step} & \multicolumn{1}{c}{5-step} & \multicolumn{1}{c}{5-step 5\%} & \multicolumn{1}{c}{avg.~steps} \\
    \cmidrule(lr){2-2} \cmidrule(lr){3-3} \cmidrule(lr){4-4} \cmidrule(lr){5-5} \cmidrule(lr){6-6} \cmidrule(lr){7-7}
    \caltech          & -0.6±0.1  & -0.1±0.0          & -0.2±0.2  & -0.0±0.1  & \textbf{-0.0±0.1} & 3.7±1.9 \\
\cifar             & inf       & inf               & 0.4±0.7   & 0.5±0.6   & \textbf{0.5±0.6}  & 2.3±1.9 \\
\cifarh            & 0.2±0.7   & 0.3±0.6           & -0.6±0.2  & 0.0±0.0   & \textbf{-0.1±1.2} & 4.7±0.5 \\
\cub              & -0.1±0.1  & -0.1±0.0          & -0.3±0.0  & -0.0±0.0  & \textbf{-0.0±0.0} & 4.3±0.9 \\
\daircraft & inf       & inf               & 13.3±11.4 & 13.3±11.4 & \textbf{0.2±0.0}  & 3.7±0.5 \\
\ddtd      & 0.7±0.4   & \textbf{0.7±0.4}  & 3.1±1.2   & 3.1±1.2   & 1.3±1.3           & 1.7±0.9 \\
\dflowers  & -0.0±0.0  & \textbf{0.0±0.0}  & 0.3±0.1   & 0.3±0.1   & 0.3±0.1           & 1.3±0.5 \\
\ducf   & 18.8±12.4 & 18.8±12.4         & 0.6±0.6   & 0.6±0.5   & \textbf{-0.1±0.1} & 4.7±0.5 \\
\eurosat             & inf       & inf               & -0.7±0.1  & 0.0±0.1   & \textbf{0.0±0.1}  & 3.3±1.2 \\
\fgvcaircrafts     & 38.8±19.4        & 38.8±19.4         & 69.6±45.6  & 69.6±45.6       & \textbf{1.0±1.0}  & 2.0±0.8 \\
\icassava            & 1.7±3.3   & 1.7±3.3    &       89.3±111.4 & 89.3±111.4       & \textbf{0.5±1.4}  & 3.7±1.9 \\
\mitd               & 0.1±0.6   & 0.3±0.4           & 0.6±1.2   & 0.8±1.1   & \textbf{0.1±0.1}  & 3.0±0.8 \\
\oxfordflowers       & 0.1±0.2   & \textbf{0.2±0.1}  & 0.1±0.2   & 0.2±0.1   & \textbf{0.2±0.1}  & 2.3±1.9 \\
\pets                & -0.8±0.0  & -0.4±0.2          & -0.8±0.0  & -0.1±0.0  & \textbf{-0.2±0.1} & 5.0±0.0 \\
\stanfordcars        & 8.4±1.8   & 8.4±1.8           & 15.5±10.2 & 15.5±10.2 & \textbf{0.2±0.2}  & 3.0±0.0 \\
\stanforddogs        & -0.3±0.4  & \textbf{-0.0±0.2} & -0.3±0.2  & -0.1±0.1  & -0.1±0.0          & 5.0±0.0                       \\
    \bottomrule
    \toprule
    \multicolumn{7}{c}{DETECTION}                                                                                                                                                                            \\
    \midrule
                        & \multicolumn{2}{c}{\textbf{powerlaw}~\cite{cortes1993learning}}                            & \multicolumn{4}{c}{\textbf{piecewise (ours)}}                                                                                            \\
    \cmidrule(lr){2-3} \cmidrule(lr){4-7}                     
                        & \multicolumn{1}{c}{1-step} & \multicolumn{1}{c}{5-step} & \multicolumn{1}{c}{1-step} & \multicolumn{1}{c}{5-step} & \multicolumn{1}{c}{5-step 5\%} & \multicolumn{1}{c}{avg.~steps} \\
    \cmidrule(lr){2-2} \cmidrule(lr){3-3} \cmidrule(lr){4-4} \cmidrule(lr){5-5} \cmidrule(lr){6-6} \cmidrule(lr){7-7}                    
    \cityscapes   & 0.4±0.8  & 0.6±0.7          & 0.1±0.4  & 0.4±0.3  & \textbf{0.4±0.3}  & 2.7±1.7 \\
\comic        & inf      & inf              & 2.4±2.7  & 2.7±2.4  & \textbf{2.7±2.5}  & 2.3±1.9 \\
\crowdhuman   & 5.5±7.2  & 5.5±7.2          & 1.0±1.5  & 1.1±1.4  & \textbf{1.1±1.4}  & 1.3±0.5 \\
\duo          & 1.7±0.7  & 1.7±0.7          & 0.8±1.0  & 0.8±0.9  & \textbf{0.1±0.2}  & 3.3±1.2 \\
\kitti        & 5.0±6.3  & 5.0±6.3          & 0.2±0.4  & 0.4±0.2  & \textbf{0.2±0.2}  & 2.3±1.9 \\
\minneapple   & -0.3±0.7 & -0.0±0.6         & -0.6±0.2 & -0.2±0.2 & \textbf{0.0±0.5}  & 4.3±0.9 \\
\sixray       & 4.0±1.9  & 4.0±1.9          & -0.2±0.2 & -0.0±0.0 & \textbf{-0.0±0.0} & 5.0±0.0 \\
\tabled       & 0.4±0.7  & 0.5±0.7          & 0.6±0.8  & 0.7±0.7  & \textbf{-0.0±0.3} & 4.3±0.9 \\
\visdrone & 0.0±0.0  & \textbf{0.0±0.0} & -0.3±0.1 & -0.1±0.0 & -0.1±0.0          & 5.0±0.0 \\
\watercolor   & 6.5±2.9  & 6.5±2.9          & 1.9±1.8  & 1.9±1.7  & \textbf{-0.1±0.4} & 4.0±1.4                      \\
    \bottomrule    
    \end{tabular}
    }
\end{minipage}
\hfill
\begin{minipage}{0.29\textwidth}
    \small
    \centering
    \includegraphics[width=1\textwidth]{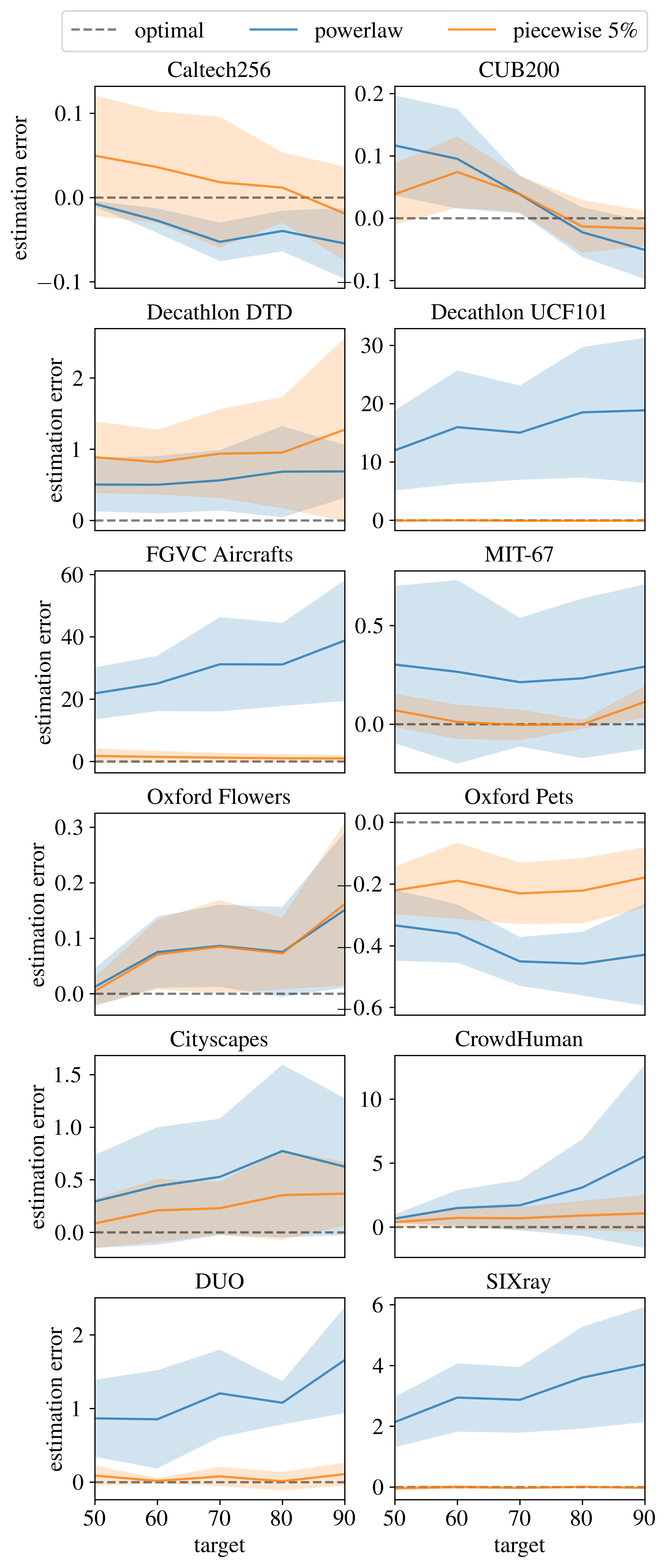}
\end{minipage}
\label{T:data_fewshot}
\end{table*}
We next evaluate the ``piecewise'' power law with the meta-model in the mid-shot setting.
We show a comparison against the ``powerlaw'' for classification datasets in Table~\ref{T:performance_midshot}.
We consider two scenarios where upto 30\% and 50\% data is used for fitting the predictors.
The meta-model used with the piecewise power law is the \textit{same} as in Section~\ref{SS:extrapolation_fewshot}.
Again, the piecewise power law is better than the power law reducing the average mean prediction errors by 50.0\% and 42.8\% when using 30\% and 50\% data for fitting, respectively, and achieving lower variance at the same time.

\subsection{Prediction of data requirements to reach target performance}
\label{SS:data_fewshot}
We provide results for maximum number steps $T \in\{1, 5\}$ and compute data estimation error $\mathcal{E}_{\mathrm{data}}$~\eqref{E:dee} which is a measure of under- or over-estimation of number of samples after $T$ steps of data collection.
We compare the ``piecewise'' power law with the meta-model against the ``powerlaw'' for $v^* = v(n_{90\%})$ for both classification and detection tasks in Table~\ref{T:data_fewshot}.
Here, $n_{90\%}$ corresponds to the subset with 90\% samples of the full dataset $\mathcal{D}^{\text{(FULL)}}$.
We also show a visualization of under-/over-estimation for different choices of the target performance corresponding to \{50, 60, 70, 80, 90\}\% data for some datasets here (and others in Appendix~\ref{A:data_fewshot}).
\begin{table*}[h!]
\rowcolors{2}{gray!10}{white}
\footnotesize
\centering
\caption{Generalization of the meta-model trained on ResNet-18 finetuning to different scenarios: ResNet-50 finetuning, ViT-B/16 finetuning, ResNet-18 linear probing, and ResNet-18 finetuning with a fixed learning rate.}
\label{T:performance_generalization}
\begin{tabular}{l|rr|rr|rr|rr}
\toprule
                    & \multicolumn{2}{c}{ResNet-50 finetune}    & \multicolumn{2}{c}{ViT-B/16 finetune} & \multicolumn{2}{c}{ResNet-18 linear} & \multicolumn{2}{c}{ResNet-18 fixed LR}    \\
                    & \textbf{powerlaw} & \textbf{piecewise} & \textbf{powerlaw}  & \textbf{piecewise}  & \textbf{powerlaw}   & \textbf{piecewise}   & \textbf{powerlaw} & \textbf{piecewise} \\
\midrule                    
\caltech          & 2.9±2.2          & \textbf{1.1±0.8}  & \textbf{1.0±0.4} &  5.1±0.4          & 4.7±3.9          & \textbf{3.2±1.5}  & 12.8±7.2 & \textbf{4.2±1.6}  \\
\cifar             & 18.9±2.2         & \textbf{1.9±1.4}  & 4.1±4.2          & \textbf{3.6±4.2}  & 15.4±0.5         & \textbf{8.3±1.3}  & 8.9±9.4  & \textbf{0.7±0.3}  \\
\cifarh            & 15.1±0.1         & \textbf{11.8±0.9} & 11.5±3.3         & \textbf{10.3±3.1} & 18.8±1.8         & \textbf{12.4±4.1} & 27.9±3.0 & \textbf{6.4±2.6}  \\
\cub              & 4.9±1.5          & \textbf{4.0±1.6}  & 6.5±2.2          & \textbf{2.6±0.4}  & 3.1±0.9          & \textbf{0.5±0.1}  & 11.9±1.1 & \textbf{1.6±0.9}  \\
\daircraft & 20.3±0.6         & \textbf{10.6±1.1} & 14.4±2.1         & \textbf{3.9±2.1}  & \textbf{2.8±0.6} & 3.5±0.6           & 13.7±1.2 & \textbf{9.9±1.4}  \\
\ddtd      & 5.6±3.0          & \textbf{5.4±1.3}  & 5.0±0.9          & \textbf{1.9±1.1}  & 1.7±0.9          & \textbf{1.3±0.0}  & 2.4±0.7  & \textbf{4.3±1.8}  \\
\dflowers  & \textbf{2.0±1.1} & 2.6±0.7           & 1.7±0.6          & \textbf{1.2±0.4}  & \textbf{0.7±0.1} & 0.8±0.2           & 3.6±0.5  & \textbf{1.4±0.1}  \\
\ducf   & 15.0±1.0         & \textbf{3.5±1.9}  & 11.5±0.3         & \textbf{6.0±3.4}  & \textbf{2.5±2.6} & 5.4±2.3           & 21.2±1.1 & \textbf{11.2±1.6} \\
\eurosat             & 4.8±4.3          & \textbf{0.3±0.1}  & 3.0±3.0          & \textbf{2.5±2.8}  & 3.6±2.4          & \textbf{3.3±0.3}  & 5.1±3.9  & \textbf{1.5±0.2}  \\
\fgvcaircrafts     & 26.6±1.8         & \textbf{15.2±5.0} & 21.8±1.1         & \textbf{15.1±2.9} & 3.8±0.7          & \textbf{2.1±1.3}  & 29.0±1.5 & \textbf{14.0±2.6} \\
\icassava            & 14.3±5.8         & \textbf{1.8±0.7}  & 11.0±3.3         & \textbf{2.1±0.7}  & 12.0±6.6         & \textbf{4.4±0.9}  & 16.4±2.3 & \textbf{6.0±3.6}  \\
\mitd               & 7.7±2.2          & \textbf{5.9±2.9}  & 5.0±2.8          & \textbf{3.2±1.5}  & 7.6±4.7          & \textbf{5.0±3.0}  & 3.4±0.4  & \textbf{3.2±0.6}  \\
\oxfordflowers       & 1.5±1.0          & \textbf{1.1±0.4}  & 0.9±0.3          & \textbf{0.5±0.1}  & 1.6±0.3          & \textbf{1.3±0.1}  & 3.6±0.1  & \textbf{2.1±0.5}  \\
\pets                & 4.1±2.8          & \textbf{2.0±0.5}  & 14.6±1.9         & \textbf{10.6±3.5} & 7.8±2.5          & \textbf{5.3±0.5}  & 10.3±0.2 & \textbf{9.0±0.4}  \\
\stanfordcars        & 23.9±1.1         & \textbf{14.7±1.4} & 20.4±0.7         & \textbf{13.9±1.4} & 7.1±0.8          & \textbf{4.4±0.9}  & 33.5±0.6 & \textbf{27.6±1.1} \\
\stanforddogs        & 8.7±3.1          & \textbf{6.8±0.5}  & \textbf{3.5±3.2} & 5.0±1.0           & 5.4±2.6          & \textbf{5.1±0.9}  & 17.9±0.1 & \textbf{11.0±1.6} \\
\midrule
AVERAGE             & 11.0±2.1         & \textbf{5.5±1.3}  & 8.5±1.9          & \textbf{5.5±1.8}  & 6.2±2.0          & \textbf{4.2±1.1}  & 13.8±2.1 & \textbf{7.1±1.3} \\
\bottomrule
\end{tabular}
\end{table*}
In the few-shot regime, the test error decays slowly in the logarithmic scale as compared to the high-shot regime.
As a result, the ``powerlaw'' over-estimates data requirements by a huge margin in several cases across both classification and detection tasks even with 1-step ($T=1$); all estimates more than 1000 are denoted as ``inf'' in Table~\ref{T:data_fewshot}.
In the cases where the power law under-estimates the performance, increasing the number of steps to $T=5$ helps reduce the data estimation error; as previously suggested in~\cite{mahmood2022much}.
We can further reduce the data estimation error by two complementary ways.
First, we improve the quality of the predictor, i.e., replace the power law by the piecewise power law.
We immediately see that ``piecewise''
reduces large overestimation by improving the quality of extrapolation.
Since the piecewise predictor is still not perfect, we further reduce the error by controlling the step sizes~\eqref{E:ppl_step} using the confidence bounds.
As a result, ``piecewise 5-step 5\%'' (referring to $\tau=5\%$) demonstrates lower data estimation error on 13/16 classification tasks and 9/10 detection tasks compared to ``powerlaw 5-step'', and consistently achieves small error with exceptions of \ddtd\ (1.3) and \comic\ (2.7). ``avg.~steps'' denotes the average number of steps taken across different random seeds with ``piecewise 5-step 5\%''.
We compute the average improvement by the PPL only on the datasets where the power law predicts less than 10$\times$ over-estimation, and yet the PPL improves the estimates by 76\% on classification and 91\% on detection datasets.

\subsection{Generalization of the meta-model}
In this section, we show generalization of the meta-model to several datasets, model architectures, and training settings.
We use the \textit{same} meta-model trained according to the procedure described in Section~\ref{SS:meta-learning} across all scenarios listed below.
Unless noted otherwise, 
we follow the few-shot setting to construct the subsets for fitting and evaluation, with the mean prediction error $\mathcal{E}_{\mathrm{perf}}$~\eqref{E:mpe} as the metric for evaluation.

\noindent\textbf{Different architectures.}
The meta-model is trained on learning curves of ResNet-18~\cite{he2015deep}.
Here, we show its generalization to a more complex architecture ResNet-50 and a different architecture ViT-B/16-224~\cite{kolesnikov2021image}.
All networks are initialized with ImageNet pretrained weights.
The piecewise power law works better than the power law on 15/16 classification tasks with ResNet-50 and 14/16 tasks with ViT-B/16; see ``ResNet-50 finetune'' and ``ViT-B/16 finetune'' in Table~\ref{T:performance_generalization}.

\noindent\textbf{Linear probing.}
The meta-model is trained on the data samples from finetuning on the subsets with ImageNet pre-trained weights.
We show that the piecewise power law with same meta-model also generalizes while linear probing on the subsets (again with ImageNet pre-trained weights) performing better than the power law on 13/16 classification tasks; see ``ResNet-18 linear'' in Table~\ref{T:performance_generalization}.

\noindent\textbf{Fixed learning rate.}
The meta-model is trained after running HPO over 3 different learning rates in \{0.001, 0.005, 0.01\} while finetuning on the subsets; see more details in Appendix~\ref{A:implementation}.
However, should it be desirable to use a fixed learning rate (we choose 0.001) during finetuning on the subsets due to limited compute, we show that the meta-model also generalizes in this scenario; see ``ResNet-18 fixed LR'' in Table~\ref{T:performance_generalization}.

\begin{table}[t!]
\rowcolors{2}{gray!10}{white}
\footnotesize
\centering
\caption{Generalization of the meta-model trained on \textit{finetuning ResNet-18} to \textit{training ResNet-18 from scratch}.}
\label{T:performance_scratch}
\vspace{-3pt}
\begin{tabular}{lrrr}
\toprule
\multicolumn{1}{c}{}  & \multicolumn{1}{c}{\textbf{powerlaw}} & \multicolumn{1}{c}{\textbf{arctan}} & \multicolumn{1}{c}{\textbf{piecewise}} \\
\midrule
\cifar~\cite{mahmood2022much}  & 39.02±20.3                            & 7.98±7.1                            & -                                      \\
\cifar \ (ours)               & 0.9±0.8                               & 2.9±0.7                             & \textbf{0.3±0.1}                       \\
\midrule
\cifarh~\cite{mahmood2022much} & 34.98±35.1                            & 13.3±5.3                            & -                                      \\
\cifarh \ (ours)              & 4.0±0.5                               & 19.4±5.3                            & \textbf{2.5±0.3}                      \\
\bottomrule
\end{tabular}
\end{table}

\begin{figure*}[t!]
\small
\centering
\includegraphics[width=1.13\columnwidth]{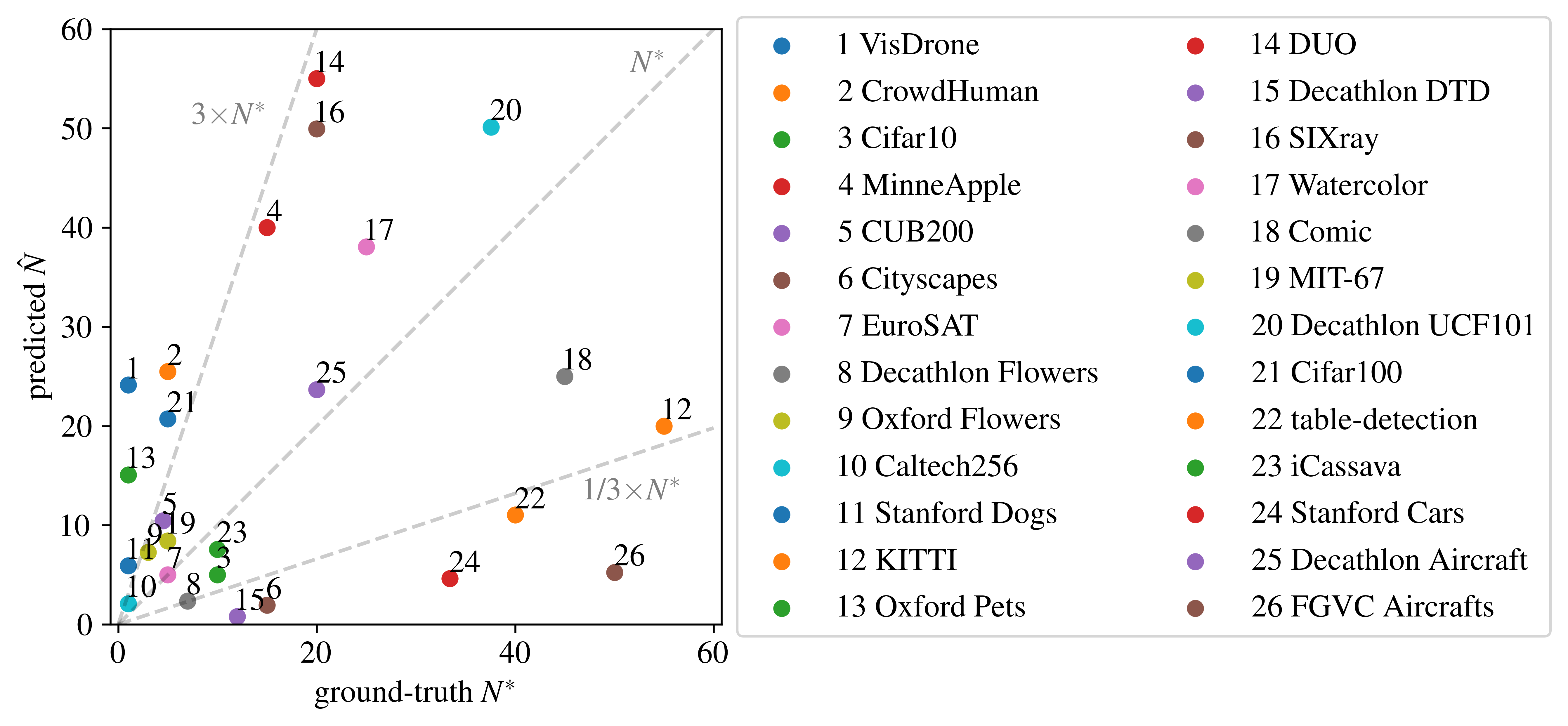}
\includegraphics[width=0.87\columnwidth]{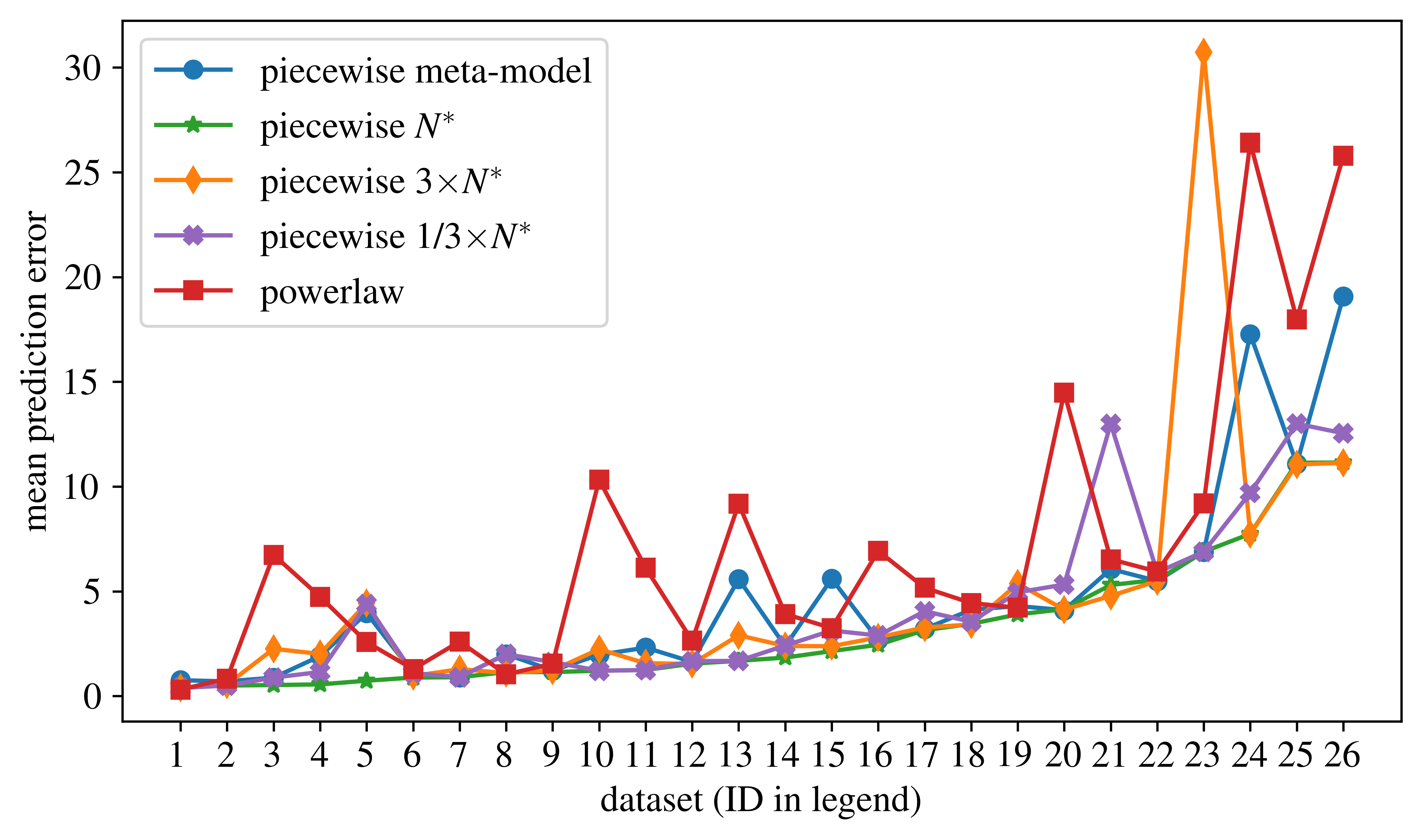}
\vspace{-5pt}
\caption{Left: Comparison of predicted versus ground-truth switch point (normalized by number of classes). Right: Mean prediction error $\mathcal{E}_{\mathrm{perf}}$~\eqref{E:mpe} obtained with powerlaw, and the piecewise power law using $N^*$, 1/3$\times N^*$, 3$\times N^*$, and the meta-model.}
\label{F:ablation_switch}
\vspace{-10pt}
\end{figure*}
\noindent\textbf{Training from scratch.}
We next show that the meta-model trained with data samples from finetuning ResNet-18 network with pre-trained models (ImageNet initialization) generalizes to training ResNet-18 network from scratch (random initialization).
In this case, we follow the settings by Mahmood et al~\cite{mahmood2022much} for constructing the subsets for fitting and evaluation of the predictors.
Specifically, we fit on subsets with \{2, 4, 6, 8, 10\}\% data and do evaluation on \{20, 30, ..., 100\}\% data using the root mean squared error (RMSE) as the metric for evaluation.
The ``piecewise'' power law with the meta-model achieves 67\% error reduction on \cifar \ and 37\% on \cifarh \  with respect to the ``powerlaw''; see Table~\ref{T:performance_scratch}.
For completeness, we also reproduce results from~\cite{mahmood2022much}; however, we note that the numbers may not be comparable due to different subsets and training recipe.
We provide a comparison with algebraic~\cite{mahmood2022much} and logarithmic~\cite{mahmood2022much} in Appendix~\ref{A:scratch}.

\subsection{Ablation studies}
\label{SS:ablation}

\noindent\textbf{Comparison of the meta-model with baselines.}
A simple baseline is to choose the switch point as the number of samples in the smallest subset, i.e., $N=n_{1}$.
We refer to it as ``linear'' baseline since the piecewise power law reduces to a linear predictor in logarithmic scale; see \eqref{E:powerlaw2}.
A second baseline is to find a ``brute-force'' solution that minimizes the error of the piecewise power law fit on the first 4 subsets $\{n_{i}, v(n_{i})\}_{i=1}^4$ and evaluated on the last subset $\{n_{5}, v(n_{5})\}$.
Note the subtle difference with respect to the procedure to compute the ground-truth $N^*$ in Section~\ref{SS:meta-learning} where the piecewise power law is fit on all 5 subsets $\{n_{i}, v(n_{i})\}_{i=1}^5$ and evaluated on the subsets in the high-shot regime.
We observe that different methods work better for different tasks but on average the ``meta-model'' works best reducing the average mean prediction error by 21.6\% and 19.1\% on ResNet-18 and ResNet-50, respectively, compared to the ``brute-force'' (next best) method.
See results in Appendix~\ref{A:baselines}.
We also note that even the ``brute-force'' method works better than the power law (see Table~\ref{T:performance_fewshot_classification} for comparison) thus demonstrating the strength of the piecewise power law.

\noindent\textbf{Quality of predictions of the meta-model.}
We compare the predictions of the meta-model against the ground-truth (GT) $N^*$ for all classification and detection tasks in Figure~\ref{F:ablation_switch} (left).
We normalize the values by the number of classes in the dataset to plot them on a similar scale.
Most predictions lie within [1/3$N^*$, 3$N^*$].
We observe that despite large errors in predictions, the meta-model performs better compared to the baselines in Table~\ref{T:performance_fewshot_classification} because the piecewise power law has high tolerance to the errors in the switch point $N$.
To demonstrate this, we evaluate two more choices of $N$ corresponding to \{1/3$N^*$, 3$N^*$\} and compare the mean prediction error in Figure~\ref{F:ablation_switch} (right) (see an elaborate comparison in Appendix~\ref{A:tolerance}).
Both of them perform better than the power law on most datasets.

\begin{figure}[th]
\small
\centering
\includegraphics[width=0.8\columnwidth]{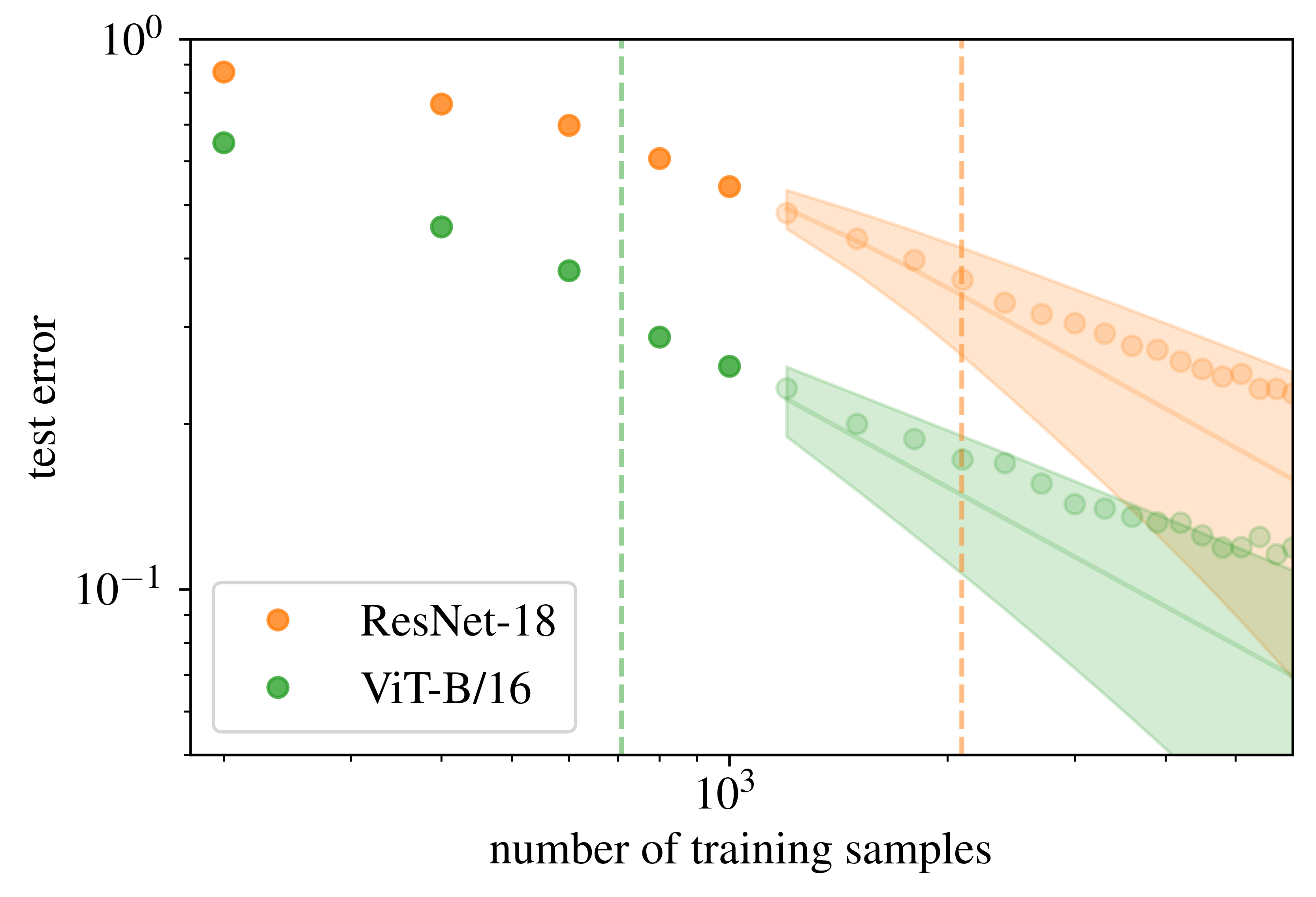}
\vspace{-5pt}
\caption{The meta-model adapts to different inputs for the same subsets finetuned with different model architectures. The piecewise power law is fit on the first 5 points (denoted in dark) and evaluated on the remaining points (denoted in light). Solid lines denote the mean prediction and the confidence bound by the shaded band. Vertical lines denote the predicted switch points.
}
\label{F:ablation_adaptability}
\vspace{-14pt}
\end{figure}

\noindent\textbf{Adaptability of the meta-model.}
For a given dataset, we expect the training statistics on the subsets to change if we finetune on the subsets differently, for example if we choose a different network architecture.
Hence the learning curves in these cases are also expected to be different.
In Figure~\ref{F:ablation_adaptability}, we show an example where the learning curve starts to exhibit linear behavior for a smaller $N$ with ViT-B/16 as compared to ResNet-18 on CUB200.
The meta-model adapts to this change and predicts a smaller switching point for ViT-B/16 as compared to ResNet-18.

\section{Conclusion}
Learning curves are better modeled by a non-linear function in the few-shot regime and a linear function in the high-shot regime.
The widely used power law best explains only the high-shot regime of the curve.
Our results show that by modeling both regimes differently with a piecewise power law (PPL) leads to lower extrapolation error on average.
Further, using the confidence bound of the PPL prevents large data estimation errors to reach a target performance.
However, the transition between the non-linear and linear regimes varies depending on the dataset, initializations, and even network architectures. 
We find that a meta learning based approach that uses statistics of the learning curves is useful to estimate the transition point.
Learning curves could exhibit other phenomena such as double descent and saturation in the very high data regimes.
We do not observe these phenomena with datasets used in this paper.
Potentially, the proposed PPL can be extended to model double descent and saturation.
We leave that for future work.

\newpage
\balance
{\small
\bibliographystyle{ieee_fullname}
\bibliography{ms}
}

\clearpage
\appendix
\balance
\section*{Appendix}
\section{Datasets}
\label{A:datasets}
\begin{table*}[ht]
\small
\rowcolors{2}{gray!10}{white}
\centering
\caption{Datasets used for experiments with the sizes of subsets used for fitting and evaluation in the few-shot regime.}
\label{T:data_stats}
\begin{tabular}{lrrrr}
\toprule
\multicolumn{5}{c}{CLASSIFICATION}                                                                                                                                                                       \\
\midrule
\rowcolor{white}
\multicolumn{1}{c}{} & \multicolumn{1}{c}{\# classes} & \multicolumn{1}{c}{\# train samples}  & \multicolumn{1}{c}{fitting}             & \multicolumn{1}{c}{evaluation}        \\
\rowcolor{white}
\multicolumn{1}{c}{} & \multicolumn{1}{c}{}           & \multicolumn{1}{c}{in largest subset} & \multicolumn{1}{c}{(samples per class)} & \multicolumn{1}{c}{(\% of full data)} \\
\midrule
\caltech           & 257                            & 15418                                 & \{1, 2, 3, 4, 5\}                       & \{10, 15, ..., 100\}                  \\
\cifar              & 10                             & 50000                                 & \{5, 10, 15, 20, 25\}                   & \{10, 15, ..., 100\}                  \\
\cifarh             & 100                            & 50000                                 & \{1, 2, 3, 4, 5\}                       & \{10, 15, ..., 100\}                  \\
\cub               & 200                            & 5994                                  & \{1, 2, 3, 4, 5\}                       & \{10, 15, ..., 100\}                  \\
\daircraft  & 100                            & 3334                                  & \{1, 2, 3, 4, 5\}                       & \{10, 15, ..., 100\}                  \\
\ddtd       & 47                             & 1880                                  & \{1, 2, 3, 4, 5\}                       & \{10, 15, ..., 100\}                  \\
\dflowers   & 102                            & 1020                                  & \{1, 2, 3, 4, 5\}                       & \{10, 15, ..., 100\}                  \\
\ducf    & 101                            & 7585                                  & \{1, 2, 3, 4, 5\}                       & \{10, 15, ..., 100\}                  \\
\eurosat              & 10                             & 20250                                 & \{5, 10, 15, 20, 25\}                   & \{10, 15, ..., 100\}                  \\
\fgvcaircrafts      & 100                            & 6667                                  & \{1, 2, 3, 4, 5\}                       & \{10, 15, ..., 100\}                  \\
\icassava             & 5                              & 4242                                  & \{10, 15, 20, 25, 30\}                  & \{10, 15, ..., 100\}                  \\
\mitd                & 67                             & 5360                                  & \{1, 2, 3, 4, 5\}                       & \{10, 15, ..., 100\}                  \\
\oxfordflowers        & 102                            & 1020                                  & \{1, 2, 3, 4, 5\}                       & \{10, 15, ..., 100\}                  \\
\pets                 & 37                             & 3680                                  & \{1, 2, 3, 4, 5\}                       & \{10, 15, ..., 100\}                  \\
\stanfordcars         & 195                            & 8144                                  & \{1, 2, 3, 4, 5\}                       & \{10, 15, ..., 100\}                  \\
\stanforddogs         & 120                            & 12000                                 & \{1, 2, 3, 4, 5\}                       & \{10, 15, ..., 100\}                 \\
\bottomrule
\toprule
\multicolumn{5}{c}{DETECTION}                                                                                                                                                                       \\
\midrule
\rowcolor{white}
\multicolumn{1}{c}{} & \multicolumn{1}{c}{\# classes} & \multicolumn{1}{c}{\# train samples}  & \multicolumn{1}{c}{fitting}             & \multicolumn{1}{c}{evaluation}          \\
\rowcolor{white}
\multicolumn{1}{c}{} & \multicolumn{1}{c}{}           & \multicolumn{1}{c}{in largest subset} & \multicolumn{1}{c}{(samples per class)} & \multicolumn{1}{c}{(samples per class)} \\
\midrule
\cityscapes           & 8                              & 800                                   & \{1, 5, 10, 15, 20\}                    & \{25, 30, ..., 100\}                    \\
\comic                & 6                              & 600                                   & \{1, 5, 10, 15, 20\}                    & \{25, 30, ..., 100\}                    \\
\crowdhuman           & 2                              & 200                                   & \{1, 5, 10, 15, 20\}                    & \{25, 30, ..., 100\}                    \\
\duo                  & 4                              & 400                                   & \{1, 5, 10, 15, 20\}                    & \{25, 30, ..., 100\}                    \\
\kitti                & 4                              & 400                                   & \{1, 5, 10, 15, 20\}                    & \{25, 30, ..., 100\}                    \\
\minneapple           & 1                              & 100                                   & \{1, 5, 10, 15, 20\}                    & \{25, 30, ..., 100\}                    \\
\sixray               & 5                              & 500                                   & \{1, 5, 10, 15, 20\}                    & \{25, 30, ..., 100\}                    \\
\tabled                & 1                              & 100                                   & \{1, 5, 10, 15, 20\}                    & \{25, 30, ..., 100\}                    \\
\visdrone         & 10                             & 1000                                  & \{1, 5, 10, 15, 20\}                    & \{25, 30, ..., 100\}                    \\
\watercolor           & 6                              & 600                                   & \{1, 5, 10, 15, 20\}                    & \{25, 30, ..., 100\}                   \\
\bottomrule
\end{tabular}
\end{table*}
We use 16 classification and 10 detection datasets for evaluation, see statistics in Table~\ref{T:data_stats}.
For classification, we randomly sample images according to subsets in Table~\ref{T:data_stats} for training and we use the original test splits for evaluation.
For detection, we choose natural k-shot sampling to construct subsets in Table~\ref{T:data_stats}
following the few-shot object detection (FSOD) setting~\cite{lee2022rethinking}.

\section{Implementation details for training models}
\label{A:implementation}
We train all models on single GPU with the following training recipe.

\noindent\textbf{Finetuning on classification datasets.}
We use ImageNet~\cite{ImageNet} pre-trained weights to initialize models for all architectures - ResNet-18~\cite{he2015deep}, ResNet-50~\cite{he2015deep}, and ViT-B/16~\cite{kolesnikov2021image}, and train for 30 epochs with a batch size of 32.
We perform HPO over 3 different learning rates (LR) $\in$ \{0.001, 0.005, 0.01\} (with the exception of ViT for which we tried a fixed LR of 0.0005) with SGD + momentum of 0.9 + weight decay of 0.0001 and LR decay by 0.1 at \{15, 25\} epochs.
We choose Top-1 Accuracy as the metric for performance $v(n$).

\noindent\textbf{Linear probing on classification datasets.}
We show experiments for ResNet-18 with ImageNet~\cite{ImageNet} pre-trained weights.
We freeze the backbone and train a linear layer with batch norm~\cite{he2022masked}.
We use a batch size of 32 and train the network for 30 epochs performing HPO over 3 different learning rates $\in$ \{0.001, 0.005, 0.01\} with SGD + momentum of 0.9 + weight decay of 0.0001 and LR decay by 0.1 at \{15, 25\} epochs.
We choose Top-1 Accuracy as the metric for performance $v(n$).

\noindent\textbf{Training from scratch on Cifar10/100.}
We show experiments for ResNet-18.
We use the hyperparameter settings from~\cite{benchopt2022} that achieves the state-of-the-art results on \cifar.
Specifically, we use a batch size of 128 and train for 100 epochs using a learning rate of 0.1 with cosine annealing~\cite{loshchilov2016sgdr} and linear warmup with SGD + momentum of 0.9 + weight decay of 0.0005.
For the full dataset, we obtained Top-1 accuracy of 0.95$\pm$0.003 on \cifar \ and 0.78$\pm$0.002 on \cifarh.

\noindent\textbf{Finetuning on detection datasets.}
We train a Faster R-CNN~\cite{ren2015faster} detector with ResNet-50+FPN backbone with COCO~\cite{COCO} pre-trained weights.
We use a batch size of $\min\{4, |\mathcal{S}_i|\}$ and train for 2000 iterations.
We choose the best learning rate $\in$ \{0.0025, 0.005\} with decay by 0.1 at 1600 iterations.
We tested the official implementation in Detectron2~\cite{wu2019detectron2} and default settings unless noted otherwise.
We choose mean Average Precision (mAP) (averaged over IoU 0.5-0.95) as the metric for performance $v(n$).

\section{Extrapolation from few-shot to high-shot}
\label{A:extrapolation_fewshot}
In Section~\ref{SS:extrapolation_fewshot}, we show the evaluation of the piecewise power law against two baselines: the power law~\cite{cortes1993learning} and arctan~\cite{mahmood2022much}.
Here, we show results for two more baselines: algebraic~\cite{mahmood2022much} and logarithmic~\cite{mahmood2022much}.
On average, the piecewise power law performs the best, followed by the power law and arctan on the classification tasks (see Table~\ref{T:performance_fewshot_classification_all}), and followed by the power law and logarithmic on the detection tasks (see Table~\ref{T:performance_fewshot_detection_all}).
\begin{table*}[ht]
\small
\rowcolors{2}{gray!10}{white}
\centering
\caption{Mean prediction error $\mathcal{E}_{\mathrm{perf}}$~\eqref{E:mpe} for extrapolating performance from few-shot to high-shot. Piecewise GT denotes the upper bound obtained with the piecewise model.}
\label{T:performance_fewshot_classification_all}
\begin{tabular}{lrrrrrr}
\toprule
\multicolumn{7}{c}{CLASSIFICATION}                                                                                                                                       \\
\midrule
                    & \multicolumn{1}{c}{\textbf{powerlaw}} & \multicolumn{1}{c}{\textbf{algebraic}} & \multicolumn{1}{c}{\textbf{arctan}}  &  \multicolumn{1}{l}{\textbf{logarithmic}} & \multicolumn{1}{c}{\textbf{piecewise}} & \multicolumn{1}{c}{\textbf{piecewise}} \\
\rowcolor{white}                    
                    & \multicolumn{1}{c}{\cite{cortes1993learning}}                  & \multicolumn{1}{c}{\cite{mahmood2022much}} & \multicolumn{1}{c}{\cite{mahmood2022much}} & \multicolumn{1}{c}{\cite{mahmood2022much}}               & \multicolumn{1}{c}{\textbf{(ours)}}    & \multicolumn{1}{c}{\textbf{(GT)}}      \\
\midrule
\caltech          & 10.3±3.6                              & 9.1±5.3            & 3.0±1.5          & 12.3±3.5             & \textbf{2.0±0.9}   & 1.2±0.5               \\
\cifar             & 6.7±2.2                               & 7.6±3.3            & 6.0±6.2          & 5.2±0.1              & \textbf{0.9±0.5}   & 0.5±0.4               \\
\cifarh            & 6.5±3.1                               & 22.5±0.1           & 17.2±7.3         & 22.2±0.4             & \textbf{6.1±3.5}   & 5.3±3.8               \\
\cub              & \textbf{2.6±0.3}                               & 18.5±0.9           & 14.1±3.9         & 16.8±1.3             & 4.0±0.1   & 0.7±0.1               \\
\daircraft & 18.0±1.8                              & 17.8±13.8          & 23.5±15.9        & 11.5±2.1             & \textbf{11.1±4.2}  & 11.1±4.1              \\
\ddtd     & 3.2±1.9                               & 5.3±1.2            & 4.7±2.5          & \textbf{3.2±0.9}     & 5.6±1.7            & 2.1±1.1               \\
\dflowers  & \textbf{1.0±0.3}                      & 1.1±0.4            & 1.5±0.3          & 1.2±0.5              & 2.0±0.3            & 1.1±0.0               \\
\ducf   & 14.5±1.9                              & 16.5±14.6          & 15.5±4.9         & 12.3±10.7            & \textbf{4.1±3.0}   & 4.1±2.9               \\
\eurosat             & 2.6±0.6                               & 4.3±3.2            & 4.2±2.4          & 2.1±0.2              & \textbf{0.9±0.2}   & 0.9±0.2               \\
\fgvcaircrafts     & 25.8±1.6                              & 18.1±2.4           & \textbf{9.2±7.4} & 10.5±6.7             & 19.1±1.4           & 11.1±1.8              \\
\icassava            & 9.2±6.7                               & 12.4±7.9           & 14.6±4.8         & 12.2±6.6             & \textbf{6.9±2.4}   & 6.9±2.4               \\
\mitd             & \textbf{4.2±1.7}                      & 15.8±6.4           & 8.2±5.3          & 11.7±6.5             & 4.3±2.5            & 3.9±2.3               \\
\oxfordflowers       & 1.5±0.4                               & 1.6±0.2            & 1.5±0.3          & 1.4±0.5              & \textbf{1.2±0.3}   & 1.1±0.4               \\
\pets                & 9.2±0.4                               & 8.4±0.7            & \textbf{1.1±0.5} & 9.7±0.2              & 5.6±0.8            & 1.7±0.5               \\
\stanfordcars        & 26.4±1.3                              & 18.8±0.4           & 17.6±2.9         & \textbf{14.2±0.6}    & 17.3±2.7           & 7.7±3.3               \\
\stanforddogs        & 6.1±5.4                               & 5.2±3.1            & 7.8±2.7          & 12.5±2.1             & \textbf{2.3±1.0}   & 1.2±0.1               \\
\midrule
AVERAGE                & 9.2±2.1                               & 11.4±4.0           & 9.3±4.3          & 9.9±2.7              & \textbf{5.8±1.6}   & 3.8±1.5               \\
\bottomrule
\end{tabular}
\end{table*}
\begin{table*}[ht]
\rowcolors{2}{gray!10}{white}
\small
\centering
\caption{Mean prediction error $\mathcal{E}_{\mathrm{perf}}$~\eqref{E:mpe} for extrapolating performance from few-shot to high-shot. Piecewise GT denotes the upper bound obtained with the piecewise model.}
\label{T:performance_fewshot_detection_all}
\begin{tabular}{lrrrrrr}
\toprule
\multicolumn{7}{c}{DETECTION}                                                                                                                                           \\
\midrule
                    & \multicolumn{1}{c}{\textbf{powerlaw}} & \multicolumn{1}{c}{\textbf{algebraic}} & \multicolumn{1}{c}{\textbf{arctan}}  &  \multicolumn{1}{c}{\textbf{logarithmic}} & \multicolumn{1}{c}{\textbf{piecewise}} & \multicolumn{1}{c}{\textbf{piecewise}} \\
\rowcolor{white}                    
                    & \multicolumn{1}{c}{\cite{cortes1993learning}}                  & \multicolumn{1}{c}{\cite{mahmood2022much}} & \multicolumn{1}{c}{\cite{mahmood2022much}} & \multicolumn{1}{c}{\cite{mahmood2022much}}               & \multicolumn{1}{c}{\textbf{(ours)}}    & \multicolumn{1}{c}{\textbf{(GT)}}      \\
\midrule
\cityscapes   & 1.3±0.8                               & 1.2±0.5            & 1.5±0.6          & 1.1±0.8              & \textbf{1.1±0.5}   & 0.9±0.6               \\
\comic        & 4.4±2.9                               & 10.0±6.3           & 28.0±33.9        & \textbf{3.9±1.7}     & 4.1±1.0            & 3.4±2.0               \\
\crowdhuman   & 0.8±0.2                               & 1.0±0.2            & 1.5±0.5          & 0.8±0.1              & \textbf{0.7±0.3}   & 0.5±0.3               \\
\duo          & 3.9±1.8                               & 5.9±4.4            & 4.5±1.6          & 2.9±2.2              & \textbf{2.4±0.5}   & 1.8±0.9               \\
\kitti        & 2.6±2.1                               & 3.3±2.0            & \textbf{1.5±1.0} & 2.2±1.1              & 1.6±0.7            & 1.5±1.4               \\
\minneapple   & 4.7±2.3                               & 1.2±0.5            & \textbf{1.1±0.3} & 1.3±0.6              & 1.9±1.0            & 0.6±0.1               \\
\sixray       & 6.9±0.9                               & 28.3±17.7          & 8.3±9.9          & 9.6±9.9              & \textbf{2.7±2.7}   & 2.4±1.1               \\
\tabled        & 5.9±2.7                               & 8.5±5.1            & 7.8±2.2          & 6.3±3.7              & \textbf{5.5±0.8}   & 5.5±2.2               \\
\visdrone & \textbf{0.3±0.1}                      & 1.0±0.4            & 0.7±0.3          & 0.9±0.3              & 0.8±0.3            & 0.4±0.1               \\
\watercolor   & 5.2±1.5                               & 19.4±22.1          & 6.7±2.7          & 6.7±3.0              & \textbf{3.2±1.4}   & 3.1±1.7               \\
\midrule
AVERAGE         & 3.6±1.5                               & 8.0±5.9            & 6.2±5.3          & 3.6±2.3              & \textbf{2.4±0.9}   & 2.0±1.0               \\
\bottomrule
\end{tabular}
\end{table*}

\section{Estimating data requirements to reach target performance}
\label{A:data_fewshot}
\begin{figure*}[t!]
\small
\centering
\includegraphics[width=0.95\textwidth]{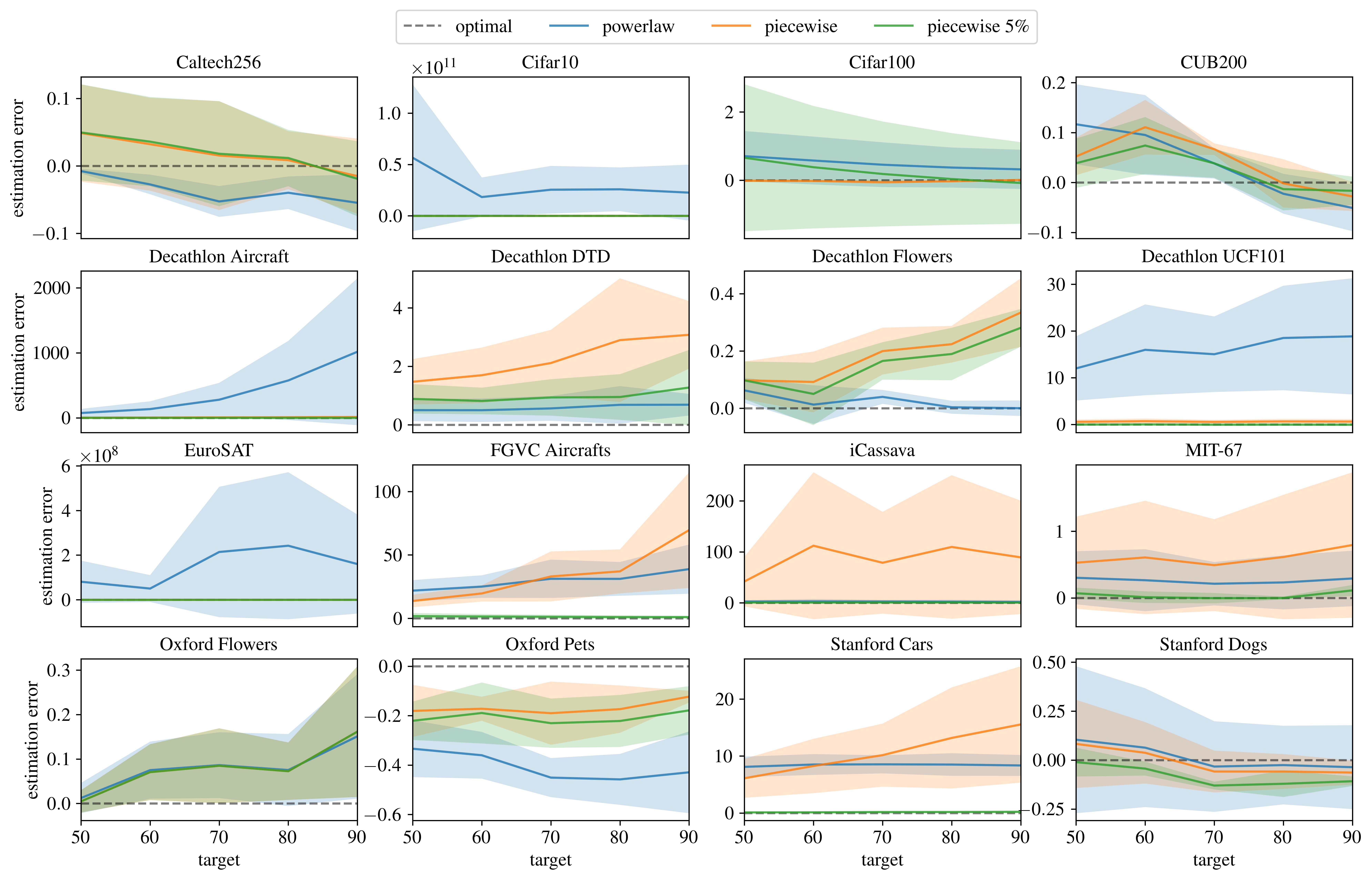}
\caption{CLASSIFICATION $T=5$: Data estimation error $\mathcal{E}_{\mathrm{data}}$~\eqref{E:dee} to reach different performance targets obtained by using \{50, 60, 70, 80, 90\}\% of the full dataset.}
\label{F:data_fewshot_classification_ablation_5}
\end{figure*}

\begin{figure*}[t!]
\small
\centering
\includegraphics[width=0.95\textwidth]{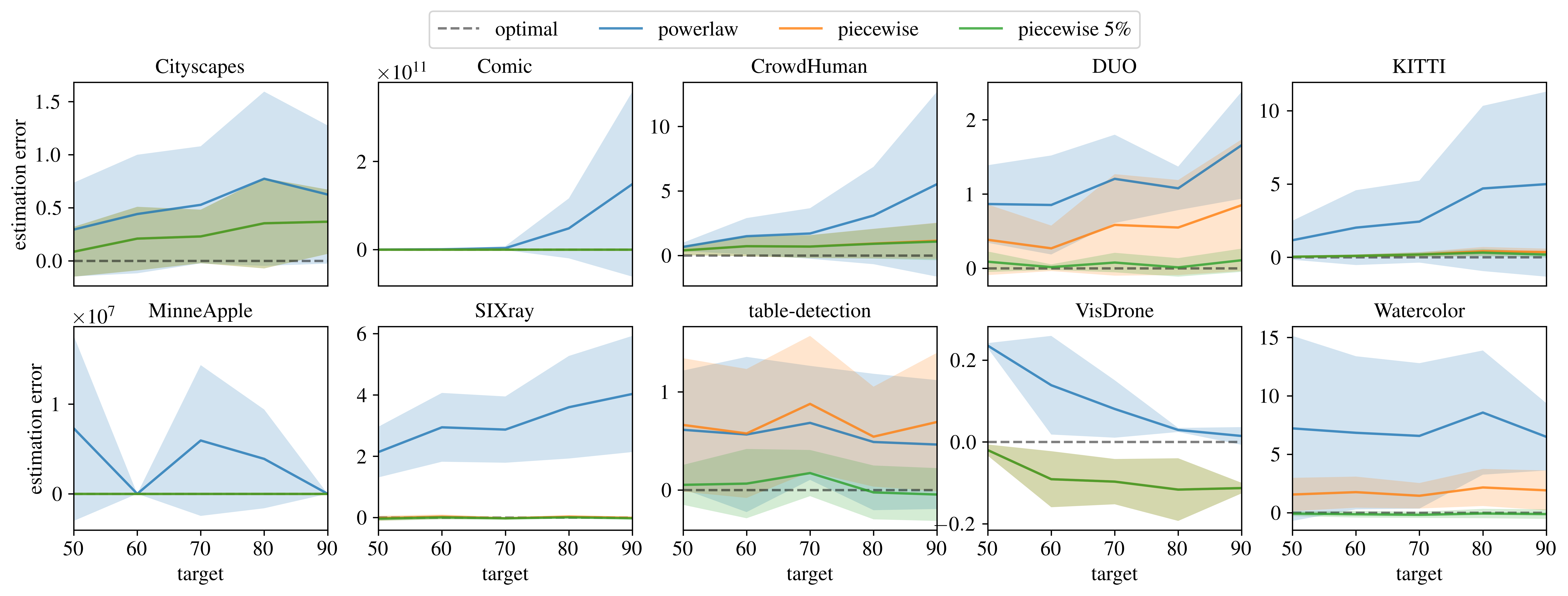}
\caption{DETECTION $T=5$: Data estimation error $\mathcal{E}_{\mathrm{data}}$~\eqref{E:dee} to reach different performance targets obtained by using \{50, 60, 70, 80, 90\}\% of the full dataset.}
\label{F:data_fewshot_detection_ablation_5}
\end{figure*}

\begin{figure*}[t!]
\small
\centering
\includegraphics[width=0.95\textwidth]{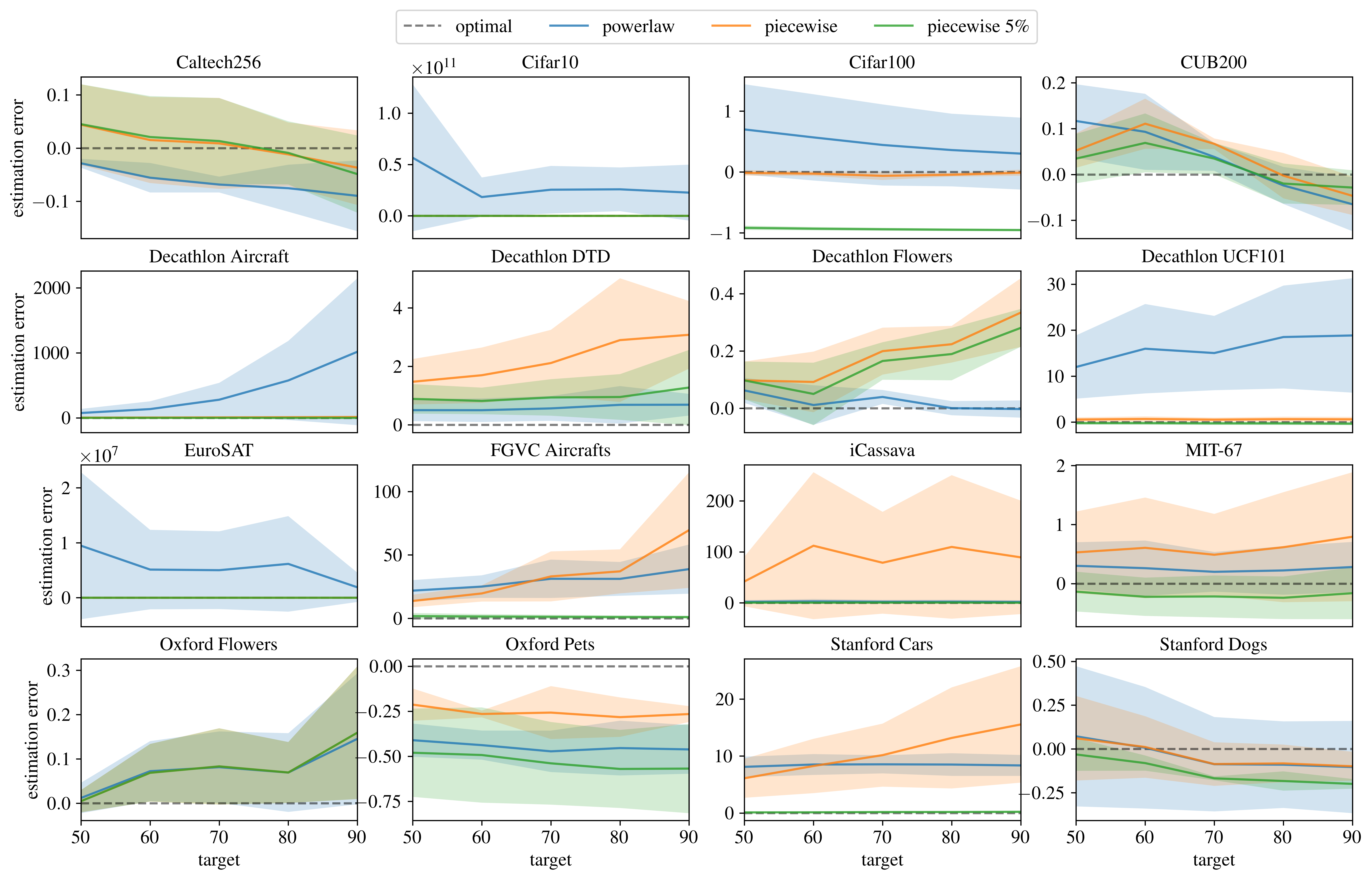}
\caption{CLASSIFICATION $T=3$: Data estimation error $\mathcal{E}_{\mathrm{data}}$~\eqref{E:dee} to reach different performance targets obtained by using \{50, 60, 70, 80, 90\}\% of the full dataset.}
\label{F:data_fewshot_classification_ablation_3}
\end{figure*}

\begin{figure*}[t!]
\small
\centering
\includegraphics[width=0.95\textwidth]{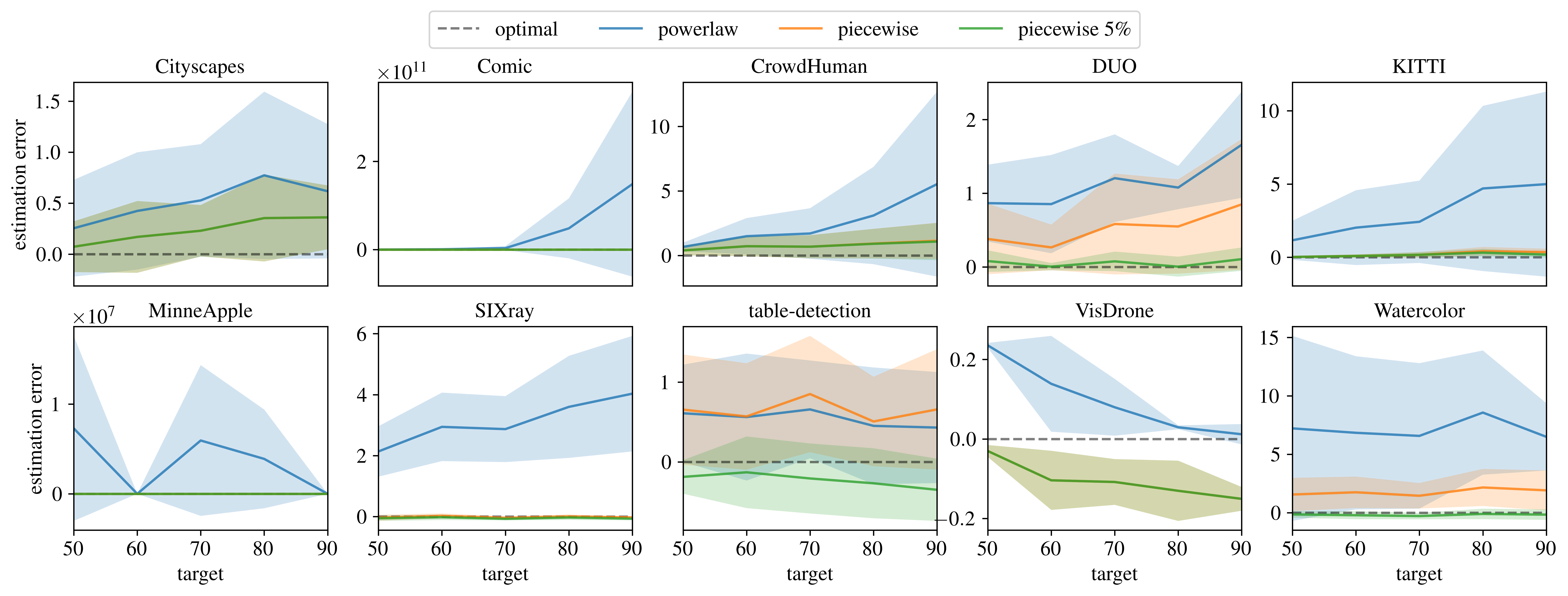}
\caption{DETECTION $T=3$: Data estimation error $\mathcal{E}_{\mathrm{data}}$~\eqref{E:dee} to reach different performance targets obtained by using \{50, 60, 70, 80, 90\}\% of the full dataset.}
\label{F:data_fewshot_detection_ablation_3}
\end{figure*}

In Section~\ref{SS:data_fewshot}, we show a visualization of data estimation error \eqref{E:dee} for different choices of the target performance corresponding to \{50, 60, 70, 80, 90\}\% data for only some datasets due to space limitation.
Here, we provide complete results for all datasets with maximum steps $T=5$ in Figure~\ref{F:data_fewshot_classification_ablation_5} and Figure~\ref{F:data_fewshot_detection_ablation_5}, and additionally with maximum steps $T=3$ in Figure~\ref{F:data_fewshot_classification_ablation_3} and Figure~\ref{F:data_fewshot_detection_ablation_3}.
The piecewise power law with ``piecewise 5\%'' (referring to $\tau=5\%$) demonstrates lower data estimation error compared to ``powerlaw'' in most cases both $T=3$ and $T=5$.

\section{Comparison with the power law}
\label{A:powerlaw_approx}
In Section~\ref{SS:ppl}, we discuss the connection between the piecewise power law~\eqref{E:ppl} and the power law~\eqref{E:powerlaw}.
Specifically, the linear term of the PPL is equivalent to the power law with its asymptotic term set to zero~\eqref{E:powerlaw2}.
We reproduce the expression here
\begin{align}
    \log(1-\hat{v}(n;\bm{\theta})) = \theta_1 + \theta_2 \log(n).
\end{align}
We refer to this predictor as ``linear'' since its parameters can simply be obtained by solving linear regression in the log-log space.
In Table~\ref{T:ablation_performance_classification}, we empirically show that ``linear'' predictor works better in mid-shot regime since the learning curve also exhibits linear behavior in the log-log space.
\begin{table*}[ht]
\rowcolors{2}{gray!10}{white}
\small
\centering
\caption{Extrapolating performance for classification tasks. Between the ``powerlaw'' and the ``linear'', we mark the better performing predictor in bold.}
\label{T:ablation_performance_classification}
\resizebox{\textwidth}{!}{
\begin{tabular}{l|rrr|rrr|rrr}
\toprule
\multicolumn{10}{c}{CLASSIFICATION}       \\
\midrule
                    & \multicolumn{3}{c}{\textbf{few-shot}}                                                                                & \multicolumn{3}{c}{\textbf{mid-shot 30\%}}                                                                           & \multicolumn{3}{c}{\textbf{mid-shot 50\%}}                                                                           \\
\rowcolor{white}                    
                    & \multicolumn{1}{c}{\textbf{powerlaw}} & \multicolumn{1}{c}{\textbf{linear}} & \multicolumn{1}{c}{\textbf{piecewise}} & \multicolumn{1}{c}{\textbf{powerlaw}} & \multicolumn{1}{c}{\textbf{linear}} & \multicolumn{1}{c}{\textbf{piecewise}} & \multicolumn{1}{c}{\textbf{powerlaw}} & \multicolumn{1}{c}{\textbf{linear}} & \multicolumn{1}{c}{\textbf{piecewise}} \\
\midrule                    
\caltech          & 10.3±3.6                              & \textbf{1.2±0.5}                    & 2.0±0.9                                & 0.8±0.6                               & \textbf{0.6±0.3}                    & 0.6±0.3                                & \textbf{0.5±0.2}                      & \textbf{0.5±0.2}                    & 0.3±0.0                                \\
\cifar             & 6.7±2.2                               & \textbf{0.9±0.5}                    & 0.9±0.5                                & 0.3±0.3                               & \textbf{0.1±0.0}                    & 0.1±0.0                                & 0.3±0.1                               & \textbf{0.1±0.0}                    & 0.1±0.0                                \\
\cifarh            & \textbf{6.5±3.1}                      & 19.0±3.0                            & 6.1±3.5                                & 1.1±0.8                               & \textbf{0.6±0.1}                    & 0.6±0.1                                & \textbf{0.3±0.1}                      & \textbf{0.3±0.1}                    & 0.3±0.1                                \\
\cub              & \textbf{2.6±0.3}                      & 8.8±0.9                             & 4.0±0.1                                & 4.5±2.4                               & \textbf{0.8±0.2}                    & 2.5±1.3                                & 1.6±0.6                               & \textbf{0.7±0.2}                    & 0.9±0.0                                \\
\daircraft & \textbf{18.0±1.8}                     & 18.5±1.7                            & 11.1±4.2                               & \textbf{8.5±1.6}                      & 10.0±1.4                            & 4.1±2.0                                & \textbf{4.2±0.3}                      & 5.9±0.2                             & 1.5±1.1                                \\
\ddtd      & \textbf{3.2±1.9}                      & 5.6±1.7                             & 5.6±1.7                                & \textbf{1.2±0.4}                      & 1.7±0.8                             & 1.7±0.8                                & 1.4±0.8                               & \textbf{1.3±0.4}                    & 1.3±0.4                                \\
\dflowers  & \textbf{1.0±0.3}                      & 3.0±0.3                             & 2.0±0.3                                & \textbf{1.4±0.4}                      & 5.9±0.8                             & 2.6±1.7                                & \textbf{1.0±0.3}                      & 3.4±0.9                             & 1.8±0.3                                \\
\ducf   & \textbf{14.5±1.9}                     & 16.1±1.6                            & 4.1±3.0                                & 2.7±1.5                               & \textbf{0.9±0.5}                    & 2.2±1.2                                & 1.0±0.4                               & \textbf{0.5±0.1}                    & 0.8±0.3                                \\
\eurosat             & 2.6±0.6                               & \textbf{0.9±0.2}                    & 0.9±0.2                                & 0.3±0.2                               & \textbf{0.1±0.0}                    & 0.1±0.0                                & \textbf{0.2±0.1}                      & 0.1±0.0                             & 0.1±0.0                                \\
\fgvcaircrafts     & \textbf{25.8±1.6}                     & 28.9±1.2                            & 19.1±1.4                               & 6.4±4.7                               & \textbf{4.0±0.5}                    & 2.0±0.8                                & 2.6±0.9                               & \textbf{2.2±0.5}                    & 0.9±0.4                                \\
\icassava            & 9.2±6.7                               & \textbf{6.9±2.4}                    & 6.9±2.4                                & 2.4±0.7                               & \textbf{1.2±0.5}                    & 1.2±0.5                                & 0.5±0.3                               & \textbf{0.5±0.1}                    & 0.5±0.1                                \\
\mitd               & \textbf{4.2±1.7}                      & 7.3±1.6                             & 4.3±2.5                                & 2.3±1.3                               & \textbf{0.8±0.2}                    & 1.1±0.4                                & 1.2±0.7                               & \textbf{0.4±0.0}                    & 0.9±0.5                                \\
\oxfordflowers       & \textbf{1.5±0.4}                      & 1.6±0.4                             & 1.2±0.3                                & 3.6±1.8                               & \textbf{3.4±0.6}                    & 1.9±0.7                                & \textbf{0.6±0.3}                      & 2.0±0.8                             & 0.7±0.5                                \\
\pets                & 9.2±0.4                               & \textbf{1.7±0.5}                    & 5.6±0.8                                & 2.2±0.9                               & \textbf{1.1±0.2}                    & 1.1±0.2                                & 1.2±0.8                               & \textbf{0.7±0.4}                    & 0.7±0.4                                \\
\stanfordcars        & \textbf{26.4±1.3}                     & 30.8±1.1                            & 17.3±2.7                               & 9.7±0.1                               & \textbf{3.4±0.1}                    & 1.1±0.4                                & 4.3±0.9                               & \textbf{1.1±0.1}                    & 0.4±0.1                                \\
\stanforddogs        & 6.1±5.4                               & \textbf{1.2±0.1}                    & 2.3±1.0                                & 3.1±2.0                               & \textbf{2.1±0.3}                    & 2.1±0.3                                & \textbf{1.2±1.2}                      & 1.6±0.3                             & 1.6±0.3                                \\
\midrule
AVERAGE             & \textbf{9.2±2.1}                      & 9.5±1.1                             & 5.8±1.6                                & 3.2±1.2                               & \textbf{2.3±0.4}                    & 1.6±0.7                                & 1.4±0.5                               & \textbf{1.3±0.3}                    & 0.8±0.3                               \\
\bottomrule
\end{tabular}
}
\end{table*}

\section{Generalization to training from scratch}
\label{A:scratch}
We provide a comparison with algebraic~\cite{mahmood2022much} and logarithmic~\cite{mahmood2022much} to show generalization of the meta-model trained on finetuning
ResNet-18 to training ResNet-18 from scratch in Table~\ref{T:performance_scratch_all}.
\begin{table*}[t!]
\rowcolors{2}{gray!10}{white}
\small
\centering
\caption{Generalization of the meta-model trained on \textit{finetuning ResNet-18} to \textit{training ResNet-18 from scratch}.}
\label{T:performance_scratch_all}
\begin{tabular}{lrrrrr}
\toprule
\multicolumn{1}{l}{}  & \textbf{powerlaw} & \textbf{algebraic} & \textbf{arctan} & \textbf{logarithmic} & \multicolumn{1}{l}{\textbf{piecewise}} \\
\rowcolor{white}                    
& \multicolumn{1}{c}{\cite{cortes1993learning}}                  & \multicolumn{1}{c}{\cite{mahmood2022much}} & \multicolumn{1}{c}{\cite{mahmood2022much}} & \multicolumn{1}{c}{\cite{mahmood2022much}}               & \multicolumn{1}{c}{\textbf{(ours)}} \\
\midrule
\cifar~\cite{mahmood2022much}  & 39.02±20.3        & 33.63±22.1         & 7.98±7.1        & 32.28±13.1           & -                             \\
\cifar \ (ours)               & 0.9±0.8           & 1.3±0.5            & 2.9±0.7         & 5.8±0.3              & \textbf{0.3±0.1}                       \\
\midrule
\cifarh~\cite{mahmood2022much} & 34.98±35.1        & 26.29±16.8         & 13.3±5.3        & 17.25±21.8           & -                                      \\
\cifarh \ (ours)              & 4.0±0.5           & 25.1±1.0           & 19.4±5.3        & 23.5±1.5             & \textbf{2.5±0.3}                      \\
\bottomrule
\end{tabular}
\end{table*}

\section{Comparison of meta-model with baselines}
\label{A:baselines}
\begin{table*}[!t]
\rowcolors{2}{gray!10}{white}
\small
\centering
\caption{Comparison of performance of the meta-model against the baselines, measured by the mean prediction error $\mathcal{E}_{\mathrm{perf}}$~\eqref{E:mpe}.}
\label{T:meta_baselines}
\begin{tabular}{l|rrr|rrr}
\toprule
\multicolumn{7}{c}{CLASSIFICATION}                                                                                                                                                                                                             \\
\midrule
\rowcolor{white}
& \multicolumn{3}{c}{\textbf{ResNet-18}}                                                                                    & \multicolumn{3}{c}{\textbf{ResNet-50}}                                                                                    \\
\rowcolor{white}
\cmidrule(lr){2-4} \cmidrule(lr){5-7}
                    & \multicolumn{1}{c}{\textbf{linear}} & \multicolumn{1}{c}{\textbf{brute-force}} & \multicolumn{1}{c}{\textbf{meta-model}} & \multicolumn{1}{c}{\textbf{linear}} & \multicolumn{1}{c}{\textbf{brute-force}} & \multicolumn{1}{c}{\textbf{meta-model}} \\
\midrule
\caltech         & \textbf{1.2±0.5} & 2.4±1.3           & 2.0±0.9                                 & \textbf{0.7±0.2} & \textbf{0.7±0.2}  & 1.1±0.8                                 \\
\cifar            & 0.9±0.5          & \textbf{0.5±0.1}  & 0.9±0.5                                 & \textbf{1.9±1.4} & 6.5±6.4           & \textbf{1.9±1.4}                        \\
\cifar            & 19.0±3.0         & \textbf{5.3±3.8}  & 6.1±3.5                                 & 7.7±0.2          & \textbf{1.2±0.3}  & 11.8±0.9                                \\
\cub              & 8.8±0.9          & \textbf{0.8±0.2}  & 4.0±0.1                                 & 6.5±1.3          & \textbf{1.9±1.2}  & 4.0±1.6                                 \\
\daircraft & 18.5±1.7         & 14.8±2.5          & \textbf{11.1±4.2}                       & 21.0±0.5         & 14.8±0.7          & \textbf{10.6±1.1}                       \\
\ddtd      & 5.6±1.7          & \textbf{4.9±0.9}  & 5.6±1.7                                 & 5.4±1.3          & \textbf{3.4±0.9}  & 5.4±1.3                                 \\
\dflowers  & 3.0±0.3          & \textbf{1.5±0.0}  & 2.0±0.3                                 & 3.0±0.4          & \textbf{2.6±0.6}  & \textbf{2.6±0.7}                        \\
\ducf   & 16.1±1.6         & 12.9±3.7          & \textbf{4.1±3.0}                        & 16.7±0.8         & 8.1±1.7           & \textbf{3.5±1.9}                        \\
\eurosat             & \textbf{0.9±0.2} & 3.0±3.3           & \textbf{0.9±0.2}                        & \textbf{0.3±0.1} & 2.8±3.3           & \textbf{0.3±0.1}                        \\
\fgvcaircrafts     & 28.9±1.2         & 20.1±2.2          & \textbf{19.1±1.4}                       & 31.3±1.2         & 20.2±2.4          & \textbf{15.2±5.0}                       \\
\icassava            & \textbf{6.9±2.4} & 25.5±18.3         & \textbf{6.9±2.4}                        & \textbf{1.8±0.7} & 19.3±24.8         & \textbf{1.8±0.7}                        \\
\mitd               & 7.3±1.6          & 4.3±3.3           & \textbf{4.3±2.5}                        & 4.0±1.4          & \textbf{2.5±2.6}  & 5.9±2.9                                 \\
\oxfordflowers       & 1.6±0.4          & \textbf{1.2±0.3}  & \textbf{1.2±0.3}                        & 1.9±0.5          & 1.2±0.4           & \textbf{1.1±0.4}                        \\
\pets                & \textbf{1.7±0.5} & 2.4±1.0           & 5.6±0.8                                 & 2.5±0.0          & 2.5±0.0           & \textbf{2.0±0.5}                        \\
\stanfordcars        & 30.8±1.1         & \textbf{16.2±2.8} & 17.3±2.7                                & 30.2±0.9         & \textbf{13.6±1.4} & 14.7±1.4                                \\
\stanforddogs        & \textbf{1.2±0.1} & 2.0±0.8           & 2.3±1.0                                 & 6.9±0.4          & 6.9±0.4           & \textbf{6.8±0.5}                        \\
\midrule
AVERAGE                & 9.5±1.1          & 7.4±2.8           & \textbf{5.8±1.6}                        & 8.9±0.7          & 6.8±3.0           & \textbf{5.5±1.3}   \\
\bottomrule
\end{tabular}
\end{table*}
In the first ablation study in Section~\ref{SS:ablation}, we compare the meta-model to two different baselines, namely (1) ``linear'' baseline (same as~\eqref{E:powerlaw2}) that uses $N=n_1$ in the piecewise power law, and (2) ``brute-force'' baseline that greedily optimizes $N$ based on the available data samples $\{n_i,v(n_i)\}_{i=1}^{5}$.
We show the results in Table~\ref{T:meta_baselines}.
We observe that different methods work better for different tasks but on average the ``meta-model'' works best reducing the average mean prediction error by 21.6\% and 19.1\% on ResNet-18 and ResNet-50, respectively, compared to the ``brute-force'' (next best) method.

\section{Quality of predictions of meta-model}
\label{A:tolerance}
\begin{table*}[t]
\rowcolors{2}{gray!10}{white}
\small
\centering
\caption{Effect on performance of choosing different switch points in the piecewise power law, measured by the mean prediction error $\mathcal{E}_{\mathrm{perf}}$~\eqref{E:mpe}.}
\label{T:ablation_switch}
\begin{tabular}{lrrrrr}
\toprule
\multicolumn{6}{c}{CLASSIFICATION}                                                                                                                                                                                                             \\
\midrule
                    & \multicolumn{1}{c}{\textbf{powerlaw}} & \multicolumn{1}{c}{\textbf{piecewise}} & \multicolumn{1}{c}{\textbf{piecewise}} & \multicolumn{1}{c}{\textbf{piecewise}} & \multicolumn{1}{c}{\textbf{piecewise}} \\
\rowcolor{white}                    
                    &  & \multicolumn{1}{c}{meta-model} & \multicolumn{1}{c}{$N^*$} & \multicolumn{1}{c}{3$\times N^*$} & \multicolumn{1}{c}{1/3$\times N^*$} \\
\midrule
\caltech          & 10.3±3.6                              & 2.0±0.9                                & 1.2±0.5                                   & 2.2±1.2                                     & 1.2±0.5                                       \\
\cifar             & 6.7±2.2                               & 0.9±0.5                                & 0.5±0.4                                   & 2.3±0.3                                     & 0.9±0.5                                       \\
\cifar            & 6.5±3.1                               & 6.1±3.5                                & 5.3±3.8                                   & 4.8±3.1                                     & 13.0±3.6                                      \\
\cub             & 2.6±0.3                               & 4.0±0.1                                & 0.7±0.1                                   & 4.4±0.2                                     & 4.4±0.8                                       \\
\daircraft & 18.0±1.8                              & 11.1±4.2                               & 11.1±4.1                                  & 11.1±4.2                                    & 13.0±3.5                                      \\
\ddtd      & 3.2±1.9                               & 5.6±1.7                                & 2.1±1.1                                   & 2.4±1.0                                     & 3.1±1.6                                       \\
\dflowers  & 1.0±0.3                               & 2.0±0.3                                & 1.1±0.0                                   & 1.1±0.0                                     & 2.0±0.3                                       \\
\ducf   & 14.5±1.9                              & 4.1±3.0                                & 4.1±2.9                                   & 4.1±2.8                                     & 5.3±3.4                                       \\
\eurosat             & 2.6±0.6                               & 0.9±0.2                                & 0.9±0.2                                   & 1.3±0.7                                     & 0.9±0.2                                       \\
\fgvcaircrafts     & 25.8±1.6                              & 19.1±1.4                               & 11.1±1.8                                  & 11.1±1.8                                    & 12.5±1.7                                      \\
\icassava            & 9.2±6.7                               & 6.9±2.4                                & 6.9±2.4                                   & 30.7±19.7                                   & 6.9±2.4                                       \\
\mitd             & 4.2±1.7                               & 4.3±2.5                                & 3.9±2.3                                   & 5.4±2.2                                     & 5.0±3.3                                       \\
\oxfordflowers       & 1.5±0.4                               & 1.2±0.3                                & 1.1±0.4                                   & 1.2±0.3                                     & 1.6±0.4                                       \\
\pets                & 9.2±0.4                               & 5.6±0.8                                & 1.7±0.5                                   & 2.9±0.6                                     & 1.7±0.5                                       \\
\stanfordcars        & 26.4±1.3                              & 17.3±2.7                               & 7.7±3.3                                   & 7.7±3.3                                     & 9.7±3.3                                       \\
\stanforddogs        & 6.1±5.4                               & 2.3±1.0                                & 1.2±0.1                                   & 1.6±0.4                                     & 1.2±0.1                                       \\
\midrule
AVERAGE             & 9.2±2.1                               & 5.8±1.6                                & 3.8±1.5                                   & 5.9±2.6                                     & 5.2±1.6                                   \\
\bottomrule
\end{tabular}
\end{table*}
We provide results to support the second ablation study in Section~\ref{SS:ablation}.
We observe that the piecewise power law has high tolerance to the errors in the switch point $N$.
To demonstrate this, we evaluate two more choices of $N$ corresponding to \{1/3$N^*$, 3$N^*$\} and compare the mean prediction error in Table~\ref{T:ablation_switch}.
Both of them perform better than the power law on most datasets.
\end{document}